\documentclass[10pt,twocolumn,letterpaper]{article}

\usepackage[pagenumbers]{cvpr} 

\usepackage[dvipsnames]{xcolor}

\usepackage{graphicx}
\usepackage{amsmath}
\usepackage{amssymb}
\usepackage{indentfirst}
\usepackage{pifont}
\usepackage{booktabs}
\usepackage{multirow}
\usepackage{algorithm}
\usepackage{algpseudocode}

\usepackage{color}
\usepackage{array}
\usepackage{colortbl}
\usepackage{cite}
\usepackage{diagbox}
\usepackage{rotating}
\usepackage{overpic}
\usepackage{contour}
\usepackage{bbding}

\newcommand{\cmark}{\ding{51}}%
\newcommand{\xmark}{\ding{55}}%
\newcommand{\tabref}[1]{Tab.~\ref{#1}}
\newcommand{\figref}[1]{Fig.~\ref{#1}}
\newcommand{\secref}[1]{Sec.~\ref{#1}}

\newcommand{\PAR}[1]{\noindent{}\textbf{#1}}

\definecolor{mygray2}{gray}{.92}



\definecolor{cvprblue}{rgb}{0.21,0.49,0.74}
\usepackage[pagebackref,breaklinks,colorlinks,citecolor=cvprblue]{hyperref}

\def\ourmodel{VPSeg}

\title{Vanishing-Point-Guided Video Semantic Segmentation of Driving Scenes}

\author{
Diandian Guo$^{1}$\quad
Deng-Ping Fan$^{3,2}$\thanks{Corresponding author: Deng-Ping Fan~\emph{(dengpfan@gmail.com)}. \\ 
Diandian Guo and Tongyu Lu share equal contributions to this project.} \quad
Tongyu Lu$^{1}$ \quad
Christos Sakaridis$^1$\quad 
Luc Van Gool$^{1}$\\
$^1$ Computer Vision Lab, ETH Z\"urich \quad 
$^2$ DISSec, CS, Nankai University \\
$^3$ Nankai International Advanced Research Institute (SHENZHEN FUTIAN) \quad \\
}

\begin{document}
\maketitle
\begin{abstract}
The estimation of implicit cross-frame correspondences and the high computational cost have long been major challenges in video semantic segmentation (VSS) for driving scenes. 
Prior works utilize keyframes, feature propagation, or cross-frame attention to address these issues.
By contrast, we are the first to harness vanishing point (VP) priors for more effective segmentation.
Intuitively, objects near VPs (\ie, away from the vehicle) are less discernible. Moreover, they tend to move radially away from the VP over time in the usual case of a forward-facing camera, a straight road, and linear forward motion of the vehicle.
Our novel, efficient network for VSS, named \textbf{\ourmodel}, incorporates two modules that utilize exactly this pair of static and dynamic VP priors: sparse-to-dense feature mining (DenseVP) and VP-guided motion fusion (MotionVP).
MotionVP employs VP-guided motion estimation to establish explicit correspondences across frames and help attend to the most relevant features from neighboring frames, while DenseVP enhances weak dynamic features in distant regions around VPs. These modules operate within a context-detail framework, which separates contextual features from high-resolution local features at different input resolutions to reduce computational costs. Contextual and local features are integrated through contextualized motion attention (CMA) for the final prediction. 
Extensive experiments on two popular driving segmentation benchmarks, Cityscapes and ACDC, demonstrate that \ourmodel~outperforms previous SOTA methods, with only modest computational overhead. The resources are available at \url{https://github.com/RascalGdd/VPSeg}.
\end{abstract}    
\section{Introduction}
\vspace{-5pt}
In the dawn of automated driving, a comprehensive understanding of the vehicle's surroundings becomes a must. Semantic segmentation, \ie, the per-pixel classification of camera frames into a set of predefined classes, is a central task in this context. However, semantic segmentation in automated driving contexts presents unique challenges. The diversity of objects, their varying scales, potential occlusions, and the wide range of lighting and weather conditions create complex visual inputs that must be parsed in real time. Especially problematic are ``hard-to-segment'' objects~\cite{fan2023advances} which are small, rare, or have an appearance that blends seamlessly into their background or into each other. These objects, such as distant traffic signs or badly lit pedestrians, are critically important to segment, given their impact on driving decisions.

\begin{figure}[t]
	\centering
	\includegraphics[scale=.12]{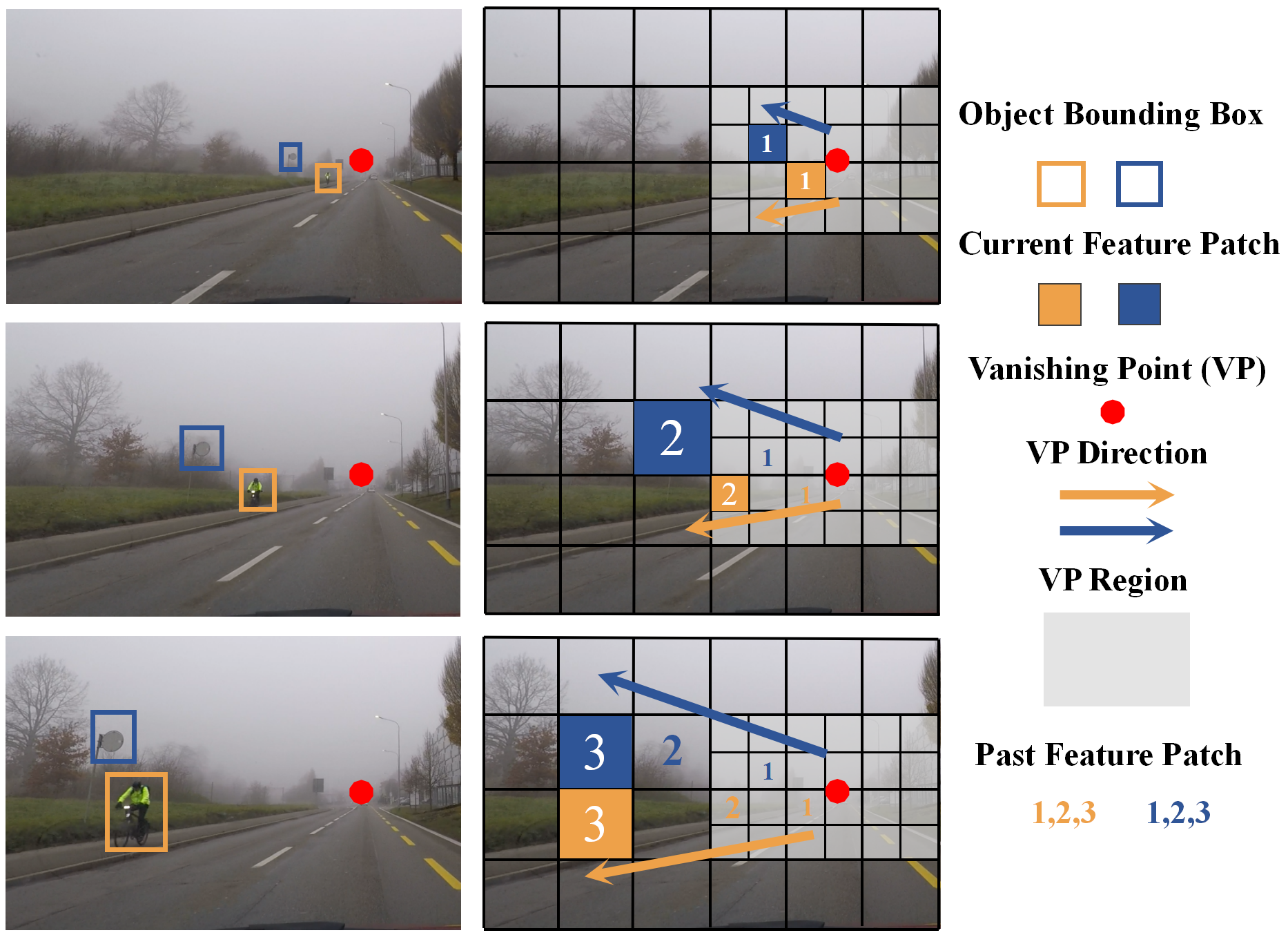}
        \vspace{-5pt} 
	\caption{\textbf{Illustration of the intuition behind our proposed vanishing-point-guided motion estimation and scale-adaptive partition modules.} Targets move radially away from the vanishing point as time progresses in the video for the typical case of a forward-facing camera, a straight road and linear forward motion, which is depicted in this example. Moreover, the region around the vanishing point contains more distant objects, which appear at smaller scales.}
	\label{intro}
\end{figure}

As the dynamic context in consecutive frames provides clues to recognize these hard samples,  researchers have tried to address this issue by exploiting temporal information through video semantic segmentation (VSS)~\cite{cffm,dff,stt,mrcfa,tcb}. However, processing multiple frames simultaneously requires significant computational resources. Moreover, existing VSS methods still struggle to establish correspondences in two main scenarios. First, for distant small objects, their relative motion over time tends to be very subtle and can easily be overlooked. Second, in high-speed driving scenes, rapid changes in object positions and appearances can pose challenges to motion estimation. Typical methods assist VSS with optical flow~\cite{accel,netwarp,dff}, which not only fails for fast motions, but also introduces higher latency. Recent works have turned to local attention to leverage temporal information~\cite{stt,cffm}. However, while the coarser granularity in the attention mechanism enables a broader context, it risks neglecting fine motion characteristics. In addition, the non-dynamic feature tracking in local attention could easily miss the fast-moving features.

Inspired by the basics of perspective projection, we hypothesize that vanishing points (VPs) can provide useful priors for addressing the above issues in VSS of driving scenes. As seen in \figref{intro}, the apparent motion of objects between consecutive frames in a video typically depends on the location of the VP, since static objects move radially away from the VP as time progresses in the usual case of a forward-facing camera, a straight road and linear forward motion. Therefore, this dynamic VP-related motion prior can serve as a constraint in motion estimation, leading to explicit cross-frame correspondences. Furthermore,the region of a frame which is located around the VP generally comprises distant parts of the driving scene, which consequently appear smaller. This static, intra-frame VP-related prior provides valuable context to quantitatively approximate and augment these crucial regions. Additionally, VP detection only requires the analysis of specific line segments or feature points in the frame, which significantly reduces the extra computational cost and does not slow down inference drastically. Thus, how to use these dynamic and static VP priors to guide VSS becomes a crucial question.

We address this question by proposing \ourmodel, a VP-guided network for VSS. \ourmodel~leverages the above dynamic and static VP priors in two respective novel modules: the VP-guided motion fusion (MotionVP) and the sparse-to-dense feature mining (DenseVP). Specifically, MotionVP establishes explicit cross-frame correspondences through VP-guided motion estimation and it thus generates the dynamic context. On the other hand, DenseVP adopts a scale-adaptive partition strategy in the region around the VP, which we refer to as ``VP region'', to extract finer features for motions in this region, which are typically indistinct. Both MotionVP and DenseVP are implemented within a context-detail framework, where dynamic and local context are extracted from bilinearly downsampled low-resolution inputs. Subsequently, we fuse the local context with the dynamic context via contextualized motion attention (CMA) to obtain the detail attention map, which guides the integration of dynamic context with high-resolution features for the final prediction.
Our contributions are summarized as:
\begin{itemize}

\item We propose MotionVP, a VP-guided motion estimation strategy for VSS, which yields explicit cross-frame correspondences. MotionVP is particularly useful in high-speed scenarios in driving scenes, with large motion.

\item We present DenseVP, a VP-guided scale-adaptive partition method for VSS, which extracts more fine-grained features for hard samples in the VP region. 

\item We design \ourmodel, an efficient context-detail framework for VSS, which adaptively separates the contextual and detail-based features with different resolutions to reduce the computational cost on video frames.

\end{itemize}

\section{Related Work}
\vspace{-5pt}

Semantic segmentation performs pixel-level labeling of an image with a set of object categories~\cite{imagesurvey}. With the advent of deep learning, various segmentation networks have been proposed~\cite{is1,is2,is3,is4,is5,is6,is7,is8,is9,is10,is11,is12,is13,is14,is15,is16} and harnessing richer \emph{spatial} context has emerged as a primary theme for this task. In contrast, video semantic segmentation (VSS) involves performing semantic segmentation on consecutive video frames and further exploits the \emph{temporal} context. Existing VSS methods can roughly be categorized into efficient VSS and high-performance VSS. While the former category compromises accuracy to speed up segmentation by reusing the features, the latter strives to enhance current frame segmentation using expensive per-frame networks. Our method falls into the second category. By guiding VSS via vanishing point priors, we achieve a better tradeoff between performance and model complexity.

\begin{figure*}[t]
	\centering
	\includegraphics[scale=.25]{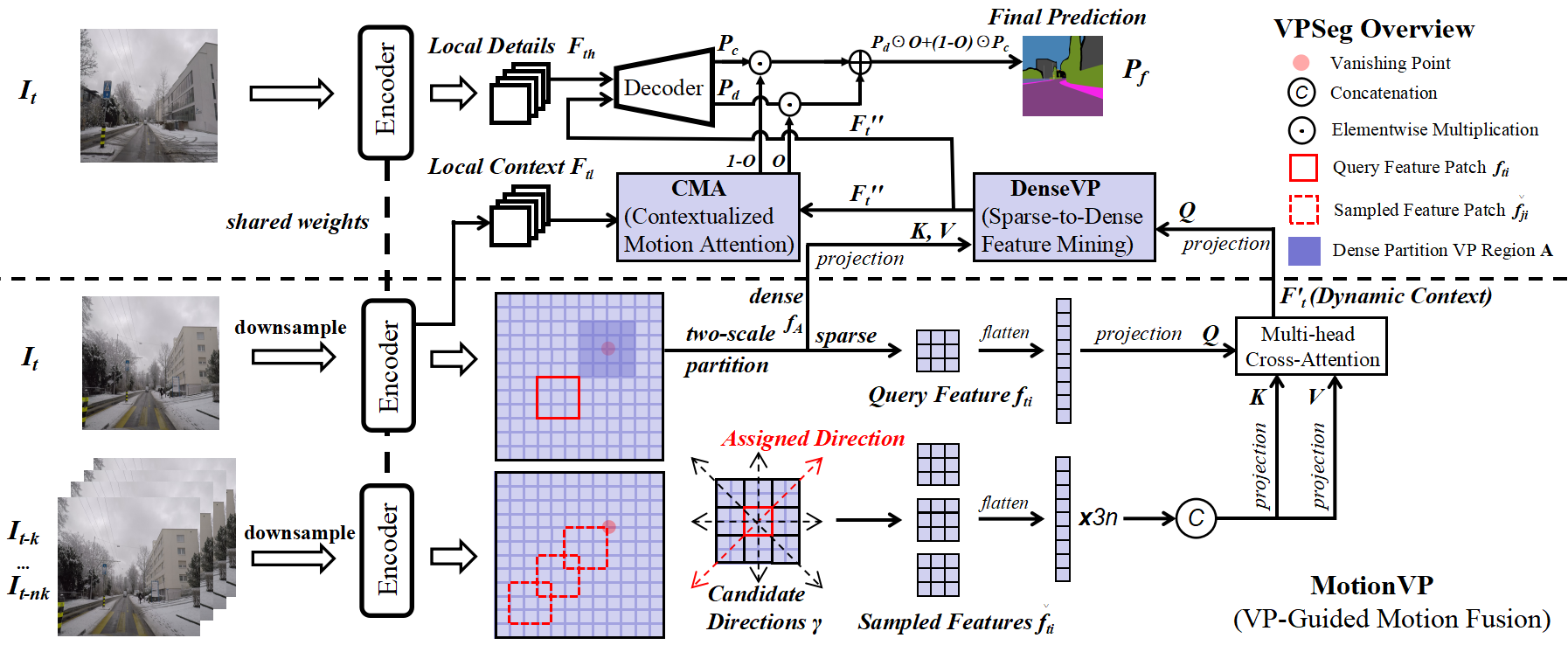}
        \vspace{-10pt}
	\caption{\textbf{Overview of our \ourmodel{} network.} In the MotionVP module (bottom part), video frames are downsampled to extract context features, which go through cross-attention to capture dynamic context ${F}'_{t}$. ${F}'_{t}$ is further augmented by DenseVP to mine finer features ${f}_{\mathbf{A}}$ in the VP region through a two-scale partition strategy. In the upper part, we obtain local context ${F}_{tl}$ and local details ${F}_{th}$ from downsampled and high-resolution target frames $I_t$, respectively. In CMA, augmented dynamic context ${F}''_{t}$ interacts with local context ${F}_{tl}$ to generate the detail attention map $O$, guiding its merging with high-resolution local details ${F}_{th}$ for the final prediction ${P}_{f}$. Zoomed in for best view.}
	\label{main}
    \vspace{-10pt}
\end{figure*}

\subsection{Efficient VSS}
\vspace{-5pt}
Efficient VSS aims to reduce the computational cost and improve the segmentation efficiency~\cite{accel,clockwork,gsvnet,tdn,lowlatency,budget,dvsnet,dff}. The most typical kind of efficient VSS approaches is the keyframe method, where the model applies expensive feature extraction and segmentation networks only on keyframes, while non-keyframe features are fine-tuned from keyframe features to reduce computation~\cite{gsvnet,lowlatency,budget,dvsnet,dff, accel}. Among these works, Accel \cite{accel} uses optical flow for feature warping \cite{netwarp}. DFF \cite{dff} propagates deep feature maps through a flow field. DVSNet \cite{dvsnet} embraces an adaptive keyframe and region scheduling policy. In addition, various other works~\cite{clockwork, tdn} also adopt the feature propagation method, where the features in preceding frames are reused to accelerate computation. Although these methods have improved the segmentation efficiency through feature sharing and propagation, their usage of proxy features often leads to inaccurate results.

\subsection{High-Performance VSS}
\vspace{-5pt}
High-performance VSS focuses on enhancing accuracy by leveraging the temporal continuity of input videos~\cite{etc, mrcfa, cffm, tcb, lma, stt}. Unlike efficient VSS approaches, these methods employ per-frame networks, utilizing a full, costly segmentation network for each frame and enhancing the current frame segmentation by mining temporal correlations from video frames. For example, Liu \etal\cite{etc} considered the temporal consistency among frames as extra constraints by using knowledge distillation for more robust VSS. Sun \etal\cite{mrcfa} estimated cross-frame affinities to enhance temporal information aggregation. Another trend in recent works \cite{cffm,lma,mrcfa} is the use of attention mechanisms \cite{attentionisallyouneed}, where the model dynamically focuses on specific parts of a video sequence to better exploit the temporal context. However, the high computational demands of these methods typically limit their application to low-resolution inputs and render them impractical when aiming for high-resolution segmentations of video frames.

\subsection{VP Detection}
\vspace{-5pt}
The vanishing point (VP) is a geometric quantity in perspective projection, which constitutes the point of apparent convergence of parallel 3D lines in the 2D image. VPs are involved in many applications, such as camera calibration \cite{vp_app1}, lane departure warning \cite{vp_app3}, and mapping \cite{vp_app4}. In contemporary research, numerous VP detection approaches have been developed.
In general terms, these methods can be divided into two categories: traditional, hand-crafted methods and deep learning-based methods. 

Traditional methods mainly include texture detection methods \cite{vp_texture1,vp_texture2,vp_texture3} and edge detection methods \cite{vp_edge1,vp_edge2}. Texture detection methods search for the dominant direction of textures in images and then vote for the VP location. They require heavy computation and cannot run in real time. In structured road scenes, Hough-transform-based~\cite{hough1} edge detection methods \cite{hough2,hough4} are more commonly used. They extract lines and transform them into Hough space to allow voting for the candidate VP. Deep learning-based methods \cite{vp_deep2,vp_deep3} primarily use CNNs \cite{cnn} to directly predict the VP location from raw image pixels. However, a lack of dedicated automated driving datasets, combined with longer inference times, hampers their practical applicability. In this work, we adopt the Hough-transform-based edge detection method \cite{hough1, hough2} to strike a balance between accuracy and inference speed.
\section{VPSeg: VP-Guided Network for VSS}

\vspace{-5pt}
\PAR{Motivation.} Structural cues such as depth maps \cite{su_depth}, layouts \cite{su_layout}, and textures \cite{su_texture} have proven essential for scene understanding. However, their application to VSS has been scarcely researched. On the one hand, such models as for depth estimation have unstable performance on distant regions. On the other hand, most deep learning-based methods are time-consuming and cumbersome, making them impractical to combine with other tasks. 

On the contrary, VP detection is swift and robust, particularly through the application of edge detection techniques across a wide range of structured driving scenes. Furthermore, motion information based on the locations of a VP in the frames of the input video can serve to capture cross-frame correspondences, given that the apparent motion of pixels across frames is often aligned with their respective offsets from this VP (cf.\ \figref{intro}). Consequently, how to exploit such VP-guided motion priors constitutes an unexplored and highly relevant question for VSS.
Notably, contemporary high-performance VSS methods are computationally demanding. Recent endeavors in this field either necessitate expensive hardware or are limited to low-resolution datasets in terms of applicability (\eg, as seen in \cite{tcb, cffm, mrcfa}). Given these findings, we pose another question: how to utilize temporal information in video frames more efficiently while maintaining high segmentation accuracy?

We found a positive answer for these two questions. Through VP-guided motion fusion (MotionVP), we establish explicit cross-frame correspondences, extracting relevant dynamic features from adjacent frames. Sparse-to-dense feature mining (DenseVP) adopts an adaptive partition of the input frame to mine finer features for subtle motions in the VP region. Additionally, our context-detail framework separates the extraction of context from detail-based features with different input resolutions. We integrate high-resolution local predictions with downsampled contextual predictions using contextualized motion attention (CMA) to reduce computation. \figref{main} illustrates the architecture of our novel \ourmodel{} network, which incorporates the above newly proposed modules.

\subsection{MotionVP: VP-Guided Motion Fusion}\label{sec:MotionVP}
\vspace{-5pt}
VSS poses the challenge of establishing cross-frame feature correspondences for the same object in frame sequences. We observe that in typical automated driving scenarios, the relative movements of static and dynamic objects normally follow the lane markings (see \figref{intro}). Typically, the driving direction coincides with the direction of the VP, which implies that most objects move radially away from the VP as time progresses in the video. 

To utilize this VP dynamic prior for motion estimation, we initialize four candidate orientations, $\gamma=n\cdot\frac{\pi}{4}, n=0,1,2,3$. Their corresponding vector representations can be denoted as $\mathcal{V} = \left \{(+1, 0), (+1, +1), (0, +1), (-1, +1) \right \}$. Assume we have a training data point containing  $n+1$ video frames $\mathcal{I}=\left \{I_{t-nk},...,I_{t-2k}, I_{t-k},I_{t} \right \} $ with corresponding timestamps $\mathcal{T}=\left \{t-nk,...,t-2k, t-k,t \right \} $, where $t-k$ is $k$ frames earlier than $t$ and $k$ is the frame sampling interval. We employ the specified $n$ previous frames to enhance the semantic segmentation of the target frame $I_t$. First, we extract context feature maps $\mathcal{F}=\left \{F_{t-nk},...,F_{t-2k}, F_{t-k},F_{t} \right \}$ from the bilinearly downsampled video frames $\mathcal{I}$ with a pre-trained transformer encoder, where $F \in \mathbb{R}^{ c\times h\times w}$ and $c,h,w$ represent channels, height and width, respectively. Then we partition the feature map into feature blocks of size $s\times s$. For the $i$-th feature patch index $(x_{i}, y_{i})$ in patch index set $\mathcal{D} =\left\{(x_i, y_i)\in \mathbb{N}^2 | x_i < \frac{w}{s}, y_i < \frac{h}{s}\right\}$, the corresponding feature patch in frame $j$ is denoted as $f_{ji} \in \mathbb{R}^{c\times s^2}$. For each patch $(x_i, y_i)$, we consider the vector pointing from the patch center to the VP as its motion direction: 
\begin{equation}\footnotesize(\Delta x_{ji}, \Delta y_{ji}) = (\hat{x}_j-x_{i}, \hat{y}_j-y_{i}),\end{equation}
where $\left ( \hat{x}_j , \hat{y}_j \right)$ is the estimated patch-level VP position in frame $j$. Different from  $(x_i, y_i) \in \mathcal{D}$, we have $(\hat{x}_j,\hat{y}_j) \in [0, \frac{w}{s}-1] \times{}[0,\frac{h}{s}-1]$. Subsequently, we mark the candidate direction that most closely matches the motion direction of the patch as its ``assigned direction''. 
The assigned direction $(u_{ji}, v_{ji})$ for patch $(x_{i}, y_{i})$ in frame $j$ can be computed as
\begin{equation}\footnotesize(u_{ji}, v_{ji}) = \underset{(u, v) \in \mathcal{V}}{\rm argmin} \  \text{dist}((u, v), (\Delta x_{ji}, \Delta y_{ji}) ), \end{equation}
where $\text{dist}(\cdot)$ quantitatively computes the difference between the motion direction and each candidate direction:
\begin{equation}\footnotesize
\text{dist}((u, v), (\Delta x, \Delta y) )=
\mid \alpha(u, v) - \alpha(\Delta x, \Delta y) \mid,
\end{equation}
where $\alpha(u, v)=$ NumPy.arctan2$(v, u)$. Then we sample adjacent patches forward and backward along the assigned direction to capture bidirectional correspondences. As the frame interval $t-j$ increases, the sampling distance $(u_{ji}, v_{ji})'$ also increases linearly with the sampling coefficient $\Delta d$, where a larger $\Delta d$ suits higher driving speeds and larger frame sampling intervals $k$:
\begin{equation}\footnotesize(u_{ji}, v_{ji})' = \frac{t-j}{k} \times \Delta d \times (u_{ji}, v_{ji}).
\end{equation}

\noindent Now, we have sampled features $\check{f}_{ji}  \in \mathbb{R}^{c\times 3s^2}$ from the concatenation of forward, backward, and local sampled patches $(\check{x}_{ji}^{f}, \check{y}_{ji}^{f})$, $(\check{x}_{ji}^{b}, \check{y}_{ji}^{b})$, and $(\check{x}_{ji}^{l}, \check{y}_{ji}^{l})$ from frame $j$:

{\footnotesize
\begin{equation}
\begin{aligned}
(\check{x}_{ji}^{f} , \check{y}_{ji}^{f}) &= (x_{i}, y_{i}) + (u_{ji}, v_{ji})', \\
(\check{x}_{ji}^{b} , \check{y}_{ji}^{b}) &= (x_{i}, y_{i}) - (u_{ji}, v_{ji})', \\
(\check{x}_{ji}^{l} , \check{y}_{ji}^{l}) &= (x_{i}, y_{i}).
\end{aligned}
\end{equation}
}

Given the local features $f_{ti} \in \mathbb{R}^{c\times s^2}$ for the $i$-th patch from the current frame $I_t$ and sampled features from neighboring frames $\mathcal{S}=\left \{ \check{f}_{ji} \in \mathbb{R}^{c\times 3s^2}, j \in \mathcal{T} \text{ and } j\ne t \right \}$, we use $f_{ti}$ as queries, $\mathcal{S}$ as keys and values for interaction:
\begin{equation}\footnotesize\label{Eq:gene_qkv}
    Q_i=\text{W}_q(f_{ti}),\quad K_i=\text{W}_k({\text{C}}(\mathcal{S})), \quad V_i=\text{W}_v({\text{C}}(\mathcal{S})),
\end{equation}
where $\text{W}(\cdot)$ and ${\text{C}}(\cdot)$ represent fully connected layers and concatenation, respectively. Next, we use cross-attention $\text{CA}(\cdot)$~\cite{attentionisallyouneed} to compute patch-level dynamic features ${f}'_{ti}\in \mathbb{R}^{c\times s^2}$ for patch $(x_{i}, y_{i})$ of frame $t$:
\begin{equation}\footnotesize\label{Eq:dynamic}
    {f}'_{ti} = \text{CA}(Q_i, K_i, V_i).
\end{equation}
The patch-level dynamic features of all patches are then simply tiled together to reconstruct the complete frame-level dynamic features ${F}'_{t}\in \mathbb{R}^{c\times h\times w}$ for frame $t$.

In summary, the VP-guided motion estimation establishes explicit feature correspondences by imposing constraints on motion estimation. This fast and simple algorithm also suits well high-speed scenarios with gradually increasing sampling distances $(u_{ji}, v_{ji})'$. It should be noted that low-resolution frames are utilized in our MotionVP module to prioritize context features over high-resolution details for more comprehensive dynamic contextual understanding. The dynamic context is afterwards fused with high-resolution details through contextualized motion attention (cf.\ \secref{sec:CMA}) to obtain the final semantic prediction.

\begin{figure*}[t!]
    \centering
    \small

    \begin{overpic}[width=.9\textwidth]{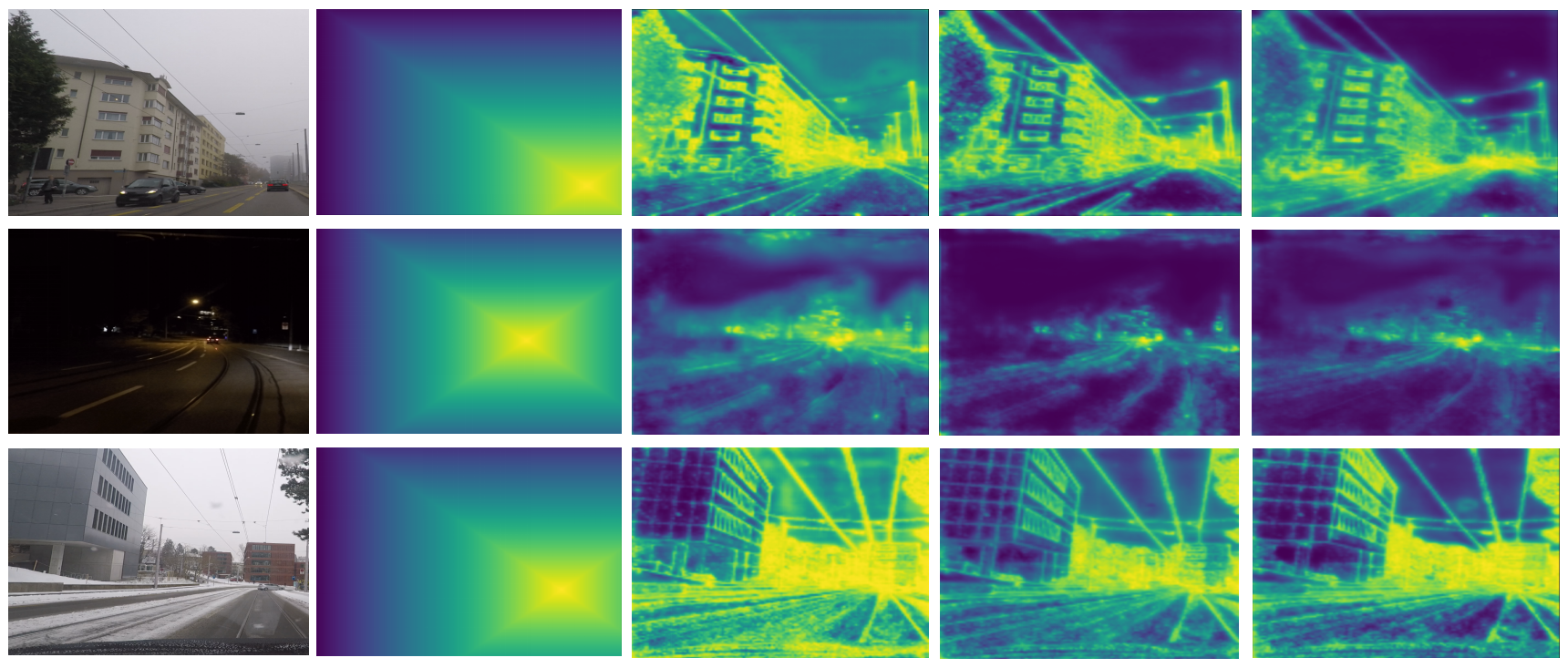} 
    \put(3,-1.5){\small (a) current frame $I_t$}
    \put(20.5,-1.5){\small (b) VP proximity map $E$}
    \put(46,-1.5){\small (c) $N=0$}
    \put(65,-1.5){\small (d) $N=1$}
    \put(86,-1.5){\small (e) $N=2$}
    \end{overpic}
    \vspace{5pt}
    \caption{\textbf{Visualization of detail attention maps $O$ with $N$ motion attention layers in CMA.} As $N$ increases, the detail attention map interacts more heavily with the dynamic features, and the weights gradually decrease in closer parts of the scene or on simple semantic categories. The highlighted distant regions near the VP suggest that the final predictions $P_f$ are primarily based on the detail-based predictions $P_{d}$ and not on $P_c$ for these regions. The VP proximity map serves as a positional prior and assists the model in pinpointing the locations of these distant regions.}\label{attn}
    \vspace{-10pt}
\end{figure*}
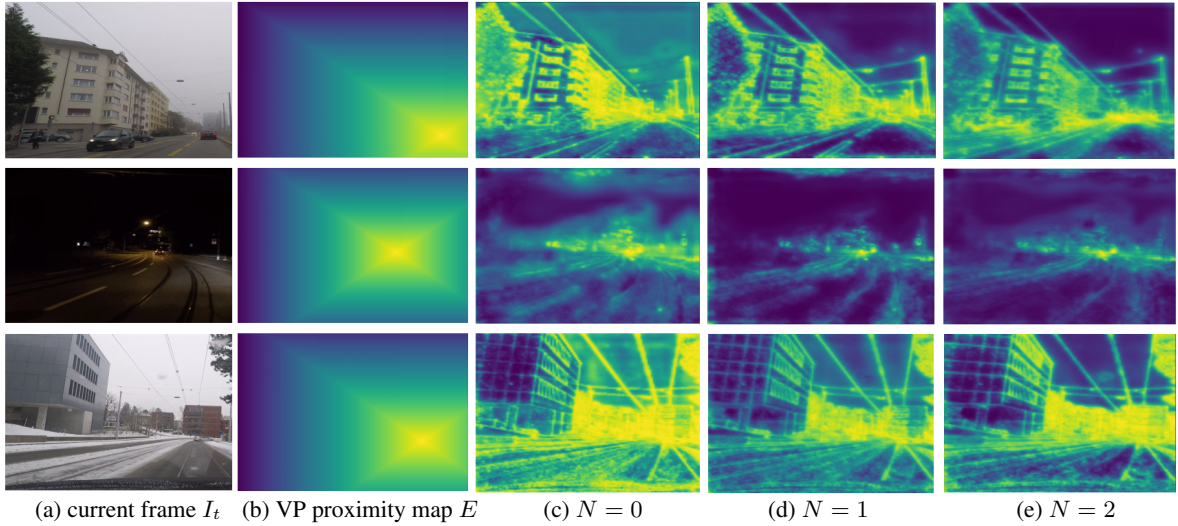

\subsection{DenseVP: Sparse-to-Dense Feature Mining}\label{sec:DenseVP}
\vspace{-5pt}
In the vicinity of the VP, objects typically appear tiny and exhibit very small motions across frames. Therefore, we propose a scale-adaptive VP-guided feature enhancement module called sparse-to-dense feature mining (DenseVP). In simple terms, we combine the dense and sparse partition to mine finer features for motions near the VP which are typically indistinct, while a sparser and more computationally efficient patch partition approach is adopted in the rest of the frame. Due to the instability in VP detection, we utilize the VP region, \ie, the region that surrounds the VP, as a coarser and more robust alternative. Specifically, we first determine the VP region based on the position of the VP. Following the notation of \secref{sec:MotionVP}, we have $\frac{h}{s} \times \frac{w}{s}$ feature patches, index $(x_i, y_i)$ for the $i$-th feature patch and estimated patch-level VP position $\left ( \hat{x}_j , \hat{y}_j \right ) $ in frame $j$.
The VP patch $(x'_{j}, y'_{j})$ represents the closest feature patch to VP and can be formulated by: 
\begin{equation}\footnotesize
    (x'_{j}, y'_{j}) = \underset{(x, y) \in \mathcal{D}}{\rm argmin} [(x-\hat{x}_j)^2+(y-\hat{y}_j)^2].
\end{equation}
The VP region $\mathcal{A} \subset \mathcal{D}$ is a rectangular region near the VP patch containing $(2a+1) \times (2b+1)$ patches arranged in a sparse grid, with $a,b \in \mathbb{N}$:
\begin{equation}\footnotesize
    \mathcal{A} = \left \{(m, n)\vert (m - x'_{j})^2 \le a^2 \ , \ (n - y'_{j})^2 \le b^2  \right \}.
\end{equation}
Then, an overlapping dense grid partition is applied in the VP region. More precisely, we adopt a stride of $\lceil s/2 \rceil$ for this dense partition, resulting in $m$ overlapping patches, where $m=(\lfloor \frac{2as}{\lceil s/2 \rceil} \rfloor+1)(\lfloor \frac{2bs}{\lceil s/2 \rceil} \rfloor+1)$. 
Subsequently, the dynamic context $F'_{t}$ is used as queries, while the features extracted densely from the VP region, $f_{\mathcal{A}} \in \mathbb{R}^{\ c \times ms^2}$, are used as keys and values:
\begin{equation}\footnotesize\label{Eq:gene_qkv}
    Q_{\mathcal{A}}=\text{W}_q(F'_{t}), \quad K_{\mathcal{A}}=\text{W}_k(f_{\mathcal{A}}), \quad V_{\mathcal{A}}=\text{W}_v(f_{\mathcal{A}}),
\end{equation}
where $\text{W}(\cdot)$ represents fully connected layers. Cross-attention operations $\text{CA}(\cdot)$ \cite{attentionisallyouneed} are then employed to augment the dynamic context $F'_{t}$ with dense features $f_{\mathcal{A}}$ for finer representations of motions in the VP region:
\begin{equation}\footnotesize\label{Eq:gene_qkv}
    F''_{t} = \text{CA}(Q_{\mathcal{A}}, K_{\mathcal{A}}, V_{\mathcal{A}}), \quad F''_{t}\in \mathbb{R}^{c\times h\times w}.
\end{equation}

The described two-scale feature partition enhances the dynamic context in the VP region, which typically contains distant objects, by leveraging static VP positional information. Thus, DenseVP exploits a static prior related to the position of the VP in a frame, while MotionVP exploits a dynamic VP prior related to apparent motion of objects depending on their relative position with respect to the VP.

\subsection{CMA: Contextualized Motion Attention}
\label{sec:CMA}
\vspace{-5pt}
Given the augmented dynamic context ${F}''_{t}\in \mathbb{R}^{c\times h\times w}$, we aim to integrate it with the high-resolution details for the final prediction. For the target frame $I_t$, the low-resolution and high-resolution inputs are denoted as $I_{tl}$ and $I_{th}$, respectively. The corresponding local context and detail-based local features are $F_{tl} \in \mathbb{R}^{ c\times h\times w}$ and $F_{th} \in \mathbb{R}^{ c\times h\times w}$. First, we randomly initialize our learnable queries $Q\in\mathbb{R}^{c \times K}$ ($K$ is the number of classes) and contextualize them with $F_{tl}$ through VP-aware cross-attention $\text{CA}_E(\cdot)$:
\begin{equation}\footnotesize
    Q_c = \text{CA}_E(\text{W}_q(Q), \text{W}_k(F_{tl}), \text{W}_v(F_{tl})),
\end{equation}
where $\text{W}(\cdot)$ denotes fully connected layers and $Q_{c} \in \mathbb{R}^{ c\times K}$ are the contextualized queries. $\text{CA}_E(\cdot)$ is given by
\begin{equation}\footnotesize
    \text{CA}_E(Q, K, V)=\text{Softmax}(\frac{QK^{T}}{\sqrt{c}}+E)V+Q.
\end{equation}
The term $E$ refers to our VP proximity map embedding. It is a  VP-centered pseudo-depth map, where the depth of pixel $(x, y)$ is $1-\Delta D$, $\Delta D \propto \max \lbrace \frac{|y-\hat{y}^p_j|}{h}, \frac{|x-\hat{x}^p_j|}{w} \rbrace$ and $(\hat{x}^p_j, \hat{y}^p_j)$ is the VP pixel coordinate. Then, we perform motion attention to merge local and dynamic context:
\begin{equation}\footnotesize
    F_m = \text{CA}_E(\text{W}_q(Q_c), \text{W}_k({F}''_{t}), \text{W}_v({F}''_{t})),
\end{equation}
\noindent where the merged context $F_{m} \in \mathbb{R}^{ c\times K}$. Thus, the detail attention map $O \in \mathbb{R}^{ K\times h\times w}$ is computed as 
\begin{equation}\footnotesize
    O = F^T_mF_{tl}.
\end{equation}
\noindent The final prediction $P_{f}$ is the weighted sum of the context prediction $P_{c}$ and the detail-based prediction $P_{d}$ under the guidance of the detail attention map $O$:
\begin{equation}\footnotesize
    P_{f} = (1-O) \odot  P_{c} + O \odot  P_{d},
\end{equation}

\noindent where $\odot$ denotes element-wise multiplication, and $P_{c}$ and $P_{d}$ are generated from $F_{tl}$ and $F_{th}$ through the DAFormer \cite{daformer} decoder. For every different semantic category and every distinct position, $O$ learns to weight the high-resolution details and dynamic context differently. As shown in \figref{attn}, the weights for $P_{d}$ are generally higher in distant regions, indicating that the final prediction $P_{f}$ in these zones are primarily determined by the detail-based prediction $P_{d}$. By contrast, in closer regions and simple semantic categories, the final results mostly rely on the context prediction $P_{c}$. Our VP proximity map serves as a positional prior embedding, guiding the detail attention map $O$ to prioritize those deeper regions. The loss function of \ourmodel~is denoted as

\begin{equation}\footnotesize
    \mathcal{L}_{\text{total}}=(1-\lambda_{d})\mathcal{L}_{\text{CE}}(P_{f}, G)+\lambda_{d}\mathcal{L}_{\text{CE}}(P_{d}, G),
\end{equation}
where $G$ and $\lambda_{d}$ represent ground truth and the loss coefficient for detail-based prediction. Here we compute the final loss as the weighted sum of the cross-entropy loss from fused and detail-based prediction, which is beneficial for improving the high-resolution details while optimizing the feature fusion.

\section{Experiments}
\vspace{-5pt}
\subsection{Experimental Setup}
\vspace{-5pt}
\PAR{Implementation details.} Our end-to-end model is trained for $160$k iterations using the AdamW \cite{adamw} optimizer. We set the batch size to $4$, with an initial learning rate of $2 \times 10^{-4}$, and the weight of the detail-based prediction loss to $\lambda_{d}=0.1$. Following the settings of SegFormer \cite{SegFormer}, we employ the MiT~\cite{SegFormer} pre-trained on ImageNet \cite{imagenet} as the backbone. For generality, only previous frames are utilized to assist segmentation for each current frame. The frame sampling interval is set to $k={3}$. We adopt random resizing, flipping, cropping, and photometric distortion for data augmentation. Unlike other high-performance VSS models \cite{tcb, cffm, mrcfa} that only use low-resolution inputs, $1024 \times 2048$ resolution inputs are employed in \ourmodel. For context features, inputs are bilinearly downsampled with a ratio of $0.5$.

For VP estimation, the thresholds of the Canny edge filter are set to $50$ and $150$ with an aperture size of $3$. The sampling coefficient $ \Delta d$ in MotionVP is $1$. We found that this number sufficiently covers fast-moving targets and achieves good performance. For DenseVP, we define the VP region as $3 \times 3$ sparse patches ($a=1, b=1$). During inference, we carry out a single-scale sliding window test. All experiments are conducted on $4$ NVIDIA RTX 3090 GPUs.

\PAR{Datasets.}
Our experiments are primarily performed on two automated driving datasets: ACDC \cite{acdc} and Cityscapes \cite{cityscapes}. ACDC consists of a large collection of images ($1,600$ clips for training and $406$ clips for validation) that are evenly distributed across four common adverse conditions: fog, nighttime, rain, and snow. More importantly, ACDC's annotation strategy involves a two-step process that creates a binary ``invalid mask'' to highlight ambiguous image regions. Initially, a semantic label is manually drafted, then refined using adverse-condition videos to finalize the annotation, allowing valid quantitative assessment of segmentation on uncertain areas. We also perform experiments on the Cityscapes video dataset, which includes 3,475 clips from 21 cities in its training and validation splits.

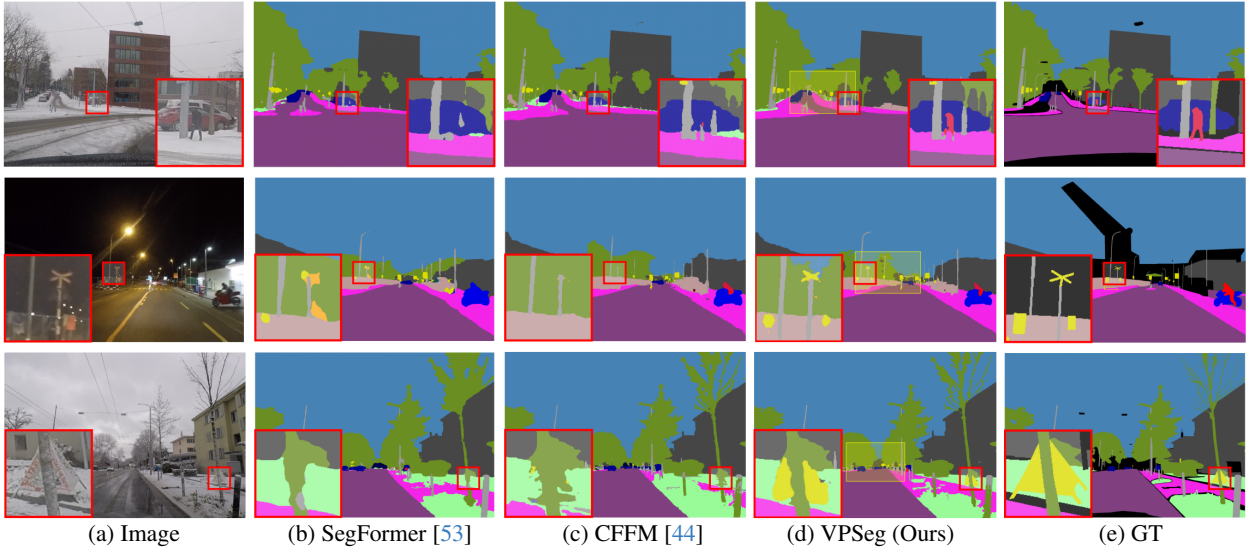
\begin{figure*}[t!]
 \centering
    \small

 \begin{overpic}[width=.95\textwidth]{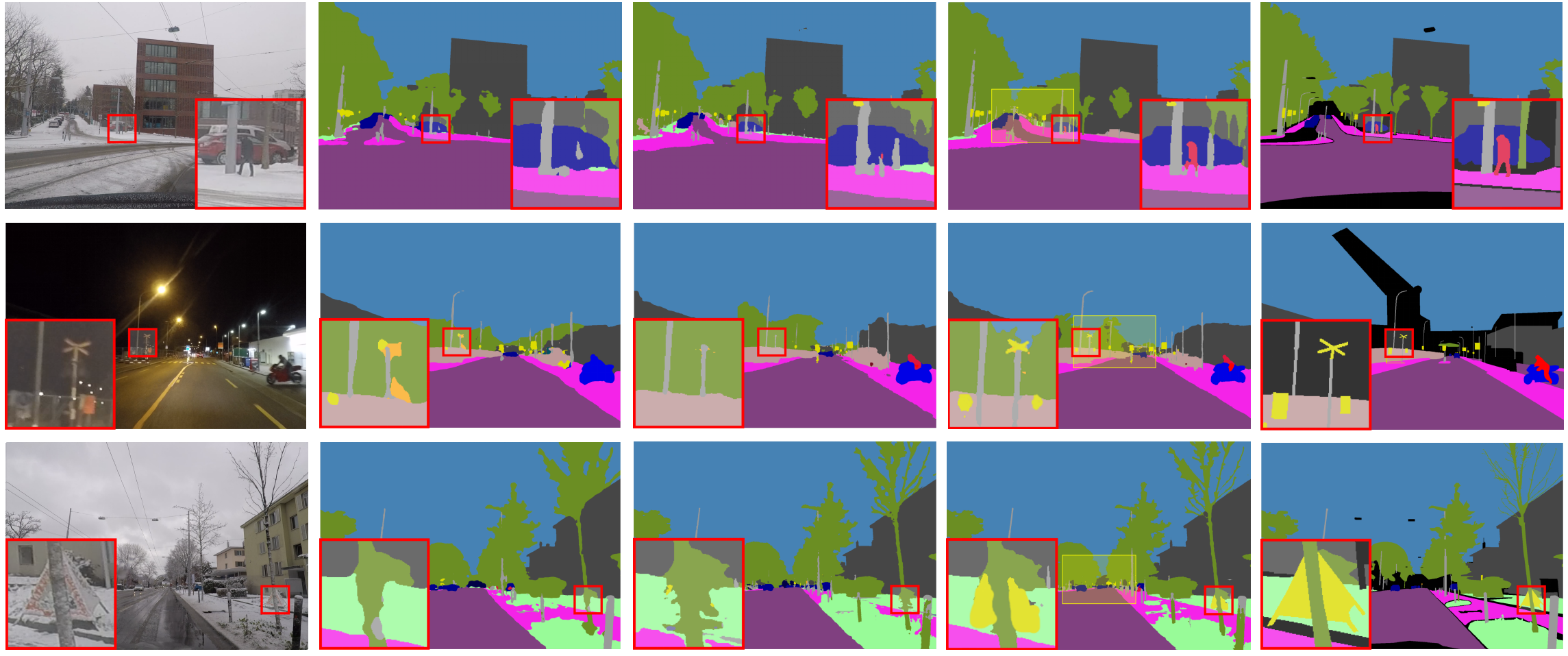} 
    \put(7,-1.5){\small (a) Image}
    \put(23,-1.5){\small (b) SegFormer \cite{SegFormer}}
    \put(45,-1.5){\small (c) CFFM \cite{cffm}}
    \put(63,-1.5){\small (d) VPSeg (Ours)}
    \put(88,-1.5){\small (e) GT}

    \end{overpic}
 \caption{\textbf{Qualitative comparison on ACDC.} The yellow box represents the densely partitioned VP region. Our model produces more accurate results for both distant tiny hard samples near the VP and occluded fast-moving close targets.}\label{visualization}
 \vspace{-10pt}
\end{figure*}

\begin{table*}[t!]
\centering
  \footnotesize
  \renewcommand{\arraystretch}{0.9}
  \setlength\tabcolsep{10.5pt}
  \begin{tabular}{r|c|c|c|c|c|c|c|c}
    \hline
    \multirow{2}{*}{Methods} & \multirow{2}{*}{Backbones} & \multirow{2}{*}{Params (M)$\downarrow$} & 
      \multicolumn{2}{c|}{mIoU$\uparrow$} &
      \multicolumn{2}{c|}{miIoU$\uparrow$} &
      {mIA-IoU$\uparrow$} & \multirow{2}{*}{VSS} \\ \cline{4-8}
      & & & {ACDC} & {Cityscapes} & {ACDC} & {Cityscapes} & {ACDC} & \\
       \hline
       
    DeepLabv3+\cite{deeplab} & ResNet-101 & 62.7 & 72.79 & 79.09 & 43.14 & 56.89 & 36.32 & \xmark\\
    
    PSPNet~\cite{psp} & ResNet-101 & 68.0 & 72.26 & 78.34 & 40.95 & 56.78 & 37.29 & \xmark\\
    
    OCRNet~\cite{ocr} & ResNet-101 & 55.6 & 70.39 & 80.09 & 43.36 & 59.55 & 33.38 & \xmark \\

    SeaFormer~\cite{seaformer} & SeaFormer-L & 14.0 & 70.09 & 77.70 & 40.41 & 56.96 & 33.26 & \xmark \\

    SegFormer~\cite{SegFormer} & MiT-B1 & \textbf{13.7} & 70.25 & 78.56 & 41.14 & 58.25 & 34.05 & \xmark \\

    SegFormer~\cite{SegFormer} & MiT-B3 & 44.6 & 75.38 & 81.32 & 46.51 & 60.01 & 38.42 & \xmark \\

    Video K-Net~\cite{videoknet} & Swin-B & 104.6 & 69.03 & 76.62 & 39.33 & 56.02 & 31.67 & \cmark \\

    ETC~\cite{etc} & ResNet-101 & 68.1 & 71.45 & 79.50 & 42.28 & 58.35 & 36.71  & \cmark \\

    TCB~\cite{tcb} & ResNet-101 & 72.5 & 70.56 & 78.70 & 41.76 & 57.84 & 35.96    & \cmark \\

    NetWarp~\cite{netwarp} & PSPNet & 90.6 & 73.71 & 80.60 & 45.59 & 59.63 & 36.58 & \cmark \\

    CFFM~\cite{cffm} & MiT-B3 & 49.6 & 75.47 & 81.44 & 47.31 & 60.09 & 37.88  & \cmark \\

    MRCFA~\cite{mrcfa} & MiT-B3 & 48.2 & 75.63 & 81.31 & 46.28 & 60.56 & 38.81 & \cmark \\ 
    \hline

    \ourmodel~(Ours) & MiT-B1 & 14.9 & 72.86 & 79.56 & 43.42 & 59.53 & 37.96 & \cmark \\

    \rowcolor{mygray2}
    \ourmodel~(Ours) & MiT-B3 & 46.8 & \textbf{77.48} & \textbf{82.46} &  \textbf{49.42} & \textbf{61.79} & \textbf{41.48}   & \cmark \\
    \hline
  \end{tabular}
    \vspace{-5pt}
    \caption{\textbf{Comparison with state-of-the-art methods on the ACDC~\cite{acdc} and Cityscapes~\cite{cityscapes} validation sets.} Our model outperforms the compared methods in mIoU, miIoU \cite{cityscapes} and mIA-IoU.}
    \label{maintable}
    \vspace{-10pt}
\end{table*}

\PAR{Evaluation metrics.}
The experiment results are evaluated on three metrics: mIoU, iIoU \cite{cityscapes}, and IA-IoU. As mIoU tends to downweigh small instances within the same class, we also employ the instance-level intersection-over-union (iIoU) for more informative evaluation with respect to performance on tiny objects. iIoU is defined as
\begin{equation}\footnotesize
    i\mathrm{IoU}=\frac{i\mathrm{TP}}{i\mathrm{TP}+i\mathrm{FN}+\mathrm{FP}},
\end{equation}
where TP, FN and FP represent true positive, false negative and false positive, respectively. The iIoU assigns higher weights $i$ to pixels in smaller instances.

For ACDC~\cite{acdc}, we propose a new metric IA-IoU (invalid-area intersection-over-union). As discussed above, the invalid masks of ACDC are particularly informative as they reveal regions of ambiguity or what might be termed as uncertain regions within the frame. Therefore, we test mIoU explicitly on these invalid regions as IA-IoU. Specifically, suppose we have $K$ semantic classes and $n_{max}$ validation images. With the corresponding invalid mask $M_{n}$ for the $n$-th image, we have:
\begin{equation}\footnotesize
    \hat {P}_{nz} = P_{nz} \cap M_{n},
\end{equation}
where $P_{nz}$ is the model prediction of the $z$-th class in the $n$-th image. Similarly, for the ground truth:
\begin{equation}\footnotesize
    \hat {G}_{nz} = G_{nz} \cap M_{n},
\end{equation}
where $G_{nz}$ is the ground truth of the $z$-th class in the $n$-th image. The IA-IoU of class $z$ can be calculated as
\begin{equation}\footnotesize
    \mathrm{IA\text{-}IoU}_z = \frac{\sum_{n=0}^{n_{max}} |\hat{P}_{nz}\cap \hat{G}_{nz}|}{\sum_{n=0}^{n_{max}}|\hat {P}_{nz}\cup \hat {G}_{nz}|}.
\end{equation}
The mean IA-IoU (mIA-IoU) is the average of IA-IoU for each class. This specialized metric allows an evaluation targeted specifically to the regions which are marked as uncertain and ambiguous in the ground truth.

\subsection{Comparison with the State of the Art}
\vspace{-5pt}

In \tabref{maintable}, we compare the performance of \ourmodel{} to state-of-the-art methods on ACDC and Cityscapes. Qualitative results are shown in \figref{visualization}. We observe that our method captures indistinct and fast motions more accurately, leading to more robust segmentation.

On ACDC, \ourmodel~improves the mIoU by $2.10$\% and $1.85$\% compared to the baseline model SegFormer~\cite{SegFormer} and the SOTA VSS model MRCFA~\cite{mrcfa}, respectively. More notable is our performance on mIA-IoU. For this metric that places more emphasis on indistinct regions, our model obtains even higher advancement, improving by $3.06$\% and $2.67$\% upon SegFormer and MRCFA, respectively. In \tabref{iaioutable}, we display the IA-IoU performance of different models for selected categories. For hard categories such as rider ($+8.5$\%) and traffic sign ($+3.4$\%), \ourmodel~delivers significant improvements. At the same time, our method matches the performance of competing methods on easier categories with larger segments, including sky ($+0.2$\%) and bus ($+0.1$\%). We also provide results on miIoU in \tabref{maintable}, where \ourmodel~performs favorably against the SOTA method CFFM \cite{cffm} (+$2.11$\%), indicating our model is better at handling distant, hard-to-segment small objects.

Besides ACDC, \ourmodel~also sets the new state of the art on Cityscapes. Compared to the baseline model SegFormer and the VSS model CFFM, our mIoU is boosted by $1.14$\% and $1.02$\%, respectively. As for miIoU, \ourmodel~improves performance even more ($+1.78$\%, $+1.70$\%), evidencing its robustness in segmenting tiny objects. Both for MiT-B1 \cite{SegFormer} and MiT-B3 \cite{SegFormer} backbones, \ourmodel~clearly outperforms the corresponding baselines, showing that the proposed modules are general.

As is shown in \tabref{flopstable}, the resource efficiency of our method is also noteworthy. In contrast to SegFormer with MiT-B3 backbone, \ourmodel~adds only $2.2$M extra parameters, which correspond to $4.9$\% of the total parameters of SegFormer ($44.6$M). Relative to other high-performance VSS models such as Video K-Net \cite{videoknet} and MRCFA, our method reaches superior computational efficiency with $1124.7$ GFLOPs. Although limited by the VP inference speed (see \tabref{table_embed}), the FPS of \ourmodel~is still comparable. Overall, \ourmodel~delivers SOTA segmentation performance with only limited additional computational cost.

\begin{table}[t!] 
  \footnotesize
  \renewcommand{\arraystretch}{.9}
  \setlength\tabcolsep{2.6pt}
    \begin{tabular}{r|cccccccccc}
    \hline
       & \rotatebox{60}{person} &  \rotatebox{60}{rider} & \rotatebox{60}{car}  & \rotatebox{60}{bus} & \rotatebox{60}{VG} & \rotatebox{60}{bicycle} & \rotatebox{60}{TS} & \rotatebox{60}{sky} & \rotatebox{60}{road}\\
    \hline
    DeepLabv3+ \cite{deeplab} & 48.2 & 18.2 & 41.3 & 40.1 & 69.6 & 26.1 & 35.0 & 84.7 & 67.7\\
    
    SegFormer \cite{SegFormer} & 47.9 & 19.0 & 44.6 & 45.9 & 71.1 & 35.4  & 32.6 & 85.4 & \textbf{74.8} \\
    
    SeaFormer \cite{seaformer} & 41.2 & 10.6 & 36.2 & 41.5 & 70.1 & 29.7  & 29.4 & 84.5 & 62.9\\
    
     ETC \cite{etc} & 45.8 & 15.5 & 41.1 & 33.5 & \textbf{71.6} & 29.6  & 34.6 & 84.5 & 70.6\\
    
    CFFM \cite{cffm} & 48.6 & 21.4 & \textbf{46.6} & 46.5 & 70.8  & 32.8 & 31.9 & 84.6 & 74.2\\
    
    MRCFA \cite{mrcfa} & 47.8 & 24.3 & 46.1 & 46.1 & 71.3 & 35.7 & 34.1 & 85.1 & 74.7\\
    \hline
    \rowcolor{mygray2}
    \ourmodel~(Ours)& \textbf{51.8} & \textbf{32.8} & 46.3 & \textbf{46.6} & 71.5 & \textbf{39.3} & \textbf{38.4}  & \textbf{85.6} & 74.5 \\
    \end{tabular}
    \vspace{-10pt}
    \caption{\textbf{Per-class IA-IoU of methods from \tabref{maintable} on ACDC.} ``VG'': vegetation, ``TS'': traffic sign. SegFormer and \ourmodel~use an MiT-B3~\cite{SegFormer} backbone.}
    \label{iaioutable}  
    \vspace{-5pt}
\end{table}

\subsection{Ablation Studies}
\vspace{-5pt}
\begin{table}[t!]
\centering
  \footnotesize
  \renewcommand{\arraystretch}{.9}
  \setlength\tabcolsep{2.0pt}
\begin{tabular}{r|c|c|c|c|c}
\hline
Methods & Backbones   & Params (M)$\downarrow$ & mIoU$\uparrow$ & GFLOPs$\downarrow$ & FPS$\uparrow$ \\ \hline
Video K-Net~\cite{videoknet}     & Swin-B & 104.6   & 69.03 & 1430.0 & -  \\ 
TMANet~\cite{tmanet}     & ResNet-101 &   54.2  & 71.62   & 1385.9 & 2.4  \\
ETC~\cite{etc}    & ResNet-101 &  68.1   & 71.45 & -  & 1.2  \\
TCB~\cite{tcb}    & ResNet-101 & 72.5   & 70.56 & -  & 1.9  \\
MRCFA~\cite{mrcfa}    & MiT-B3 &  48.2  & 75.63  & 1436.4   &  \textbf{4.4}  \\ 
CFFM~\cite{cffm}     & MiT-B3 & 49.6  & 75.47  & 1534.8 & 3.8\\ 
\hline
\rowcolor{mygray2}
\ourmodel~(Ours)    & MiT-B3 &   \textbf{46.8}    & \textbf{77.48}  & \textbf{1124.7} & 3.4 \\ 
\end{tabular}
\vspace{-10pt}
\caption{Comparison of FPS and GFLOPs of high-performance VSS methods on ACDC.}
\label{flopstable}
\vspace{-5pt}
\end{table}

\noindent\textbf{Benefit of DenseVP.}\quad
In \tabref{table_area}, we conduct ablation experiments on the sparse-to-dense feature mining (DenseVP) module under different settings. As the VP region expands, the performance of the model gradually improves and peaks when $3 \times 3$ ($a=1, b=1$) feature patches are used. IA-IoU is more significantly influenced by DenseVP than mIoU, indicating that the scale-adaptive partition in DenseVP contributes more to the enhancement of uncertain regions.
\begin{table}[t!]
\centering
  \footnotesize
  \renewcommand{\arraystretch}{.9}
  \setlength\tabcolsep{8pt}
\begin{tabular}{r|c|c|c}
\hline
 VP region size & mIoU (A.)$\uparrow$ & mIA-IoU (A.)$\uparrow$ & mIoU (C.)$\uparrow$    \\ \hline 
  No DenseVP    &   77.15           & 41.02  & 82.19    \\
  $a=0, b=0$    &   77.28           & 40.87  & 82.34  \\
  \rowcolor{mygray2}
  $a=1, b=1$    &   \textbf{77.48}  & \textbf{41.48}   & 82.46  \\ 
  $a=2, b=2$    &   77.36           & 41.33   & 82.37  \\
  $a=3, b=3$    &   77.41           & 41.32   & \textbf{82.50}  \\

\end{tabular}
\vspace{-10pt}
\caption{Ablation study on the VP region size on ACDC (A.) and Cityscapes (C.) with MiT-B3 backbone.}
\label{table_area}
\vspace{-5pt}
\end{table}

\begin{table}[t!]
\centering
\footnotesize
  \renewcommand{\arraystretch}{.9}
  \setlength\tabcolsep{3.2pt}
\begin{tabular}{r|c|c|c|c}
\hline
 Embeddings  & mIoU (A.)$\uparrow$ & mIA-IoU (A.)$\uparrow$ & mIoU (C.)$\uparrow$  & FPS$\uparrow$ \\ \hline
  No embedding    &    77.02   & 40.91  & 81.16  & \textbf{4.2} \\
  Depth map   &    77.19   & 40.78  & 82.23  &  1.8 \\
  \rowcolor{mygray2}
  VP proximity map   & \textbf{77.48}  & \textbf{41.48}  & \textbf{82.46} & 3.4 \\
  Depth map + VP   &    77.46   & 41.37   & 82.39 &  1.6  \\ 

\end{tabular}
\vspace{-10pt}
\caption{Ablation study of different positional embeddings on ACDC (A.) and Cityscapes (C.) with MiT-B3 backbone.}
\label{table_embed}
\vspace{-10pt}
\end{table}

\noindent\textbf{Effect of different positional embeddings.}\quad
Given that VP proximity maps are constructed as pseudo depth maps, we attempt to replace the VP proximity embedding in CMA with depth map embeddings. The experimental results are presented in \tabref{table_embed}, where the depth maps are generated by MiDaS V3-Hybrid \cite{midas}. As the depth estimation has unstable performances at distant regions, the depth map embedding yields limited improvement in mIoU and leads to worse performance on IA-IoU. In addition, the longer inference time of MiDaS results in a significant drop in the FPS. The VP proximity embedding alone reaches the best mIoU and mIA-IoU with a medium inference speed. Experiments incorporating both depth map and VP proximity embeddings have comparable results, but still limited FPS.

\noindent\textbf{Effect of the frame sampling interval $k$.}\quad
We study the impact of the frame sampling interval $k$ in \tabref{table_distance}. Notably, \ourmodel~obtains even higher mIoU when $k$ increases from $3$ to $5$. For larger $k$, \ourmodel~also exhibits more robustness against other methods whose mIoU drops drastically. The possible reason is that the axial motion searching along the VP direction covers a wider range of motions, which is especially relevant for high-speed scenarios.

\noindent\textbf{Influence of the number of motion attention layers $N$.}\quad
\tabref{table_layer} shows the performance of \ourmodel~with respect to the number of motion attention layers $N$ in CMA. As $N$ increases, the model performance significantly improves, proving the effectiveness of our dynamic feature fusion. We observe diminishing returns in performance for $N\ge 2$. For \ourmodel~with MiT-B3 backbone, we choose $N=2$ for the best tradeoff between performance and model complexity.

\begin{table}[t!]
\centering
  \footnotesize
  \renewcommand{\arraystretch}{.9}
  \setlength\tabcolsep{10.2pt}
\begin{tabular}{r|c|c|c|c}
\hline
 Methods~/~$k$   & 3 & 5 & 7 & 9   \\ \hline
  Segformer \cite{SegFormer}    &  75.38   & 75.38  & 75.38 & 75.38   \\
  MRCFA \cite{mrcfa}  &    75.63  & 75.44  & 75.32 & 75.30  \\
  CFFM \cite{cffm} &   75.47  & 75.35  & 75.22 & 75.07 \\
  \hline
  \rowcolor{mygray2}  
  \ourmodel~(Ours)    &  ~\textbf{77.48 } & ~\textbf{77.52 }   & \textbf{77.39} & ~\textbf{77.43 } \\ 
  
\end{tabular}
\vspace{-10pt}
\caption{Ablation study of the frame sampling interval $k$ (default $3$) on ACDC with MiT-B3 backbone.}
\label{table_distance}
\vspace{-5pt}
\end{table}

\begin{table}[t!]
\centering
\footnotesize
  \renewcommand{\arraystretch}{.9}
  \setlength\tabcolsep{5.8pt}
\begin{tabular}{r|c|c|c|c}
\hline
  $N$  & mIoU (A.)$\uparrow$  & mIA-IoU (A.)$\uparrow$ & mIoU (C.)$\uparrow$  & Params (M)$\downarrow$   \\ \hline
  $0$    &   76.65  & 40.33  & 82.12  & \textbf{45.9} \\
  $1$   &    77.21  & 41.23  & 82.33 & 46.4 \\
  \rowcolor{mygray2}
  $2$    & 77.48  & \textbf{41.48}  & \textbf{82.46} & 46.8 \\
  $3$   &   \textbf{77.51}   & 41.39   & 82.44 & 47.3 \\ 

\end{tabular}
\vspace{-10pt}
\caption{Ablation study on the number of motion attention layers in CMA with MiT-B3 backbone.}
\label{table_layer}
\vspace{-10pt}
\end{table}

\section{Conclusion}
\vspace{-5pt}
The establishment of cross-frame correspondences and the reduction of computational cost are two pressing issues in VSS of driving scenes. We have proposed the novel \ourmodel{} network to address these issues by exploiting dynamic and static VP priors through the novel MotionVP and DenseVP modules. The former module establishes explicit correspondences via a VP-guided motion estimation strategy, while the latter augments fine dynamic features in the region around the estimated VP through a scale-adaptive partition method. In the context-detail framework of \ourmodel{}, the downsampled contextual features and high-resolution local details are separated and adaptively fused through our motion-aware CMA attention module. \ourmodel{} has achieved SOTA performance for VSS on two widely used driving datasets at a reasonable computational cost.

\section{Acknowledgments}
This work is funded by Toyota Motor Europe via the research project TRACE-Z\"urich.

{
    \small
    \bibliographystyle{ieeenat_fullname}
    \bibliography{main}
}

\newpage
\appendix

\section{Vanishing Point Detection}

We adopt a classical solution for vanishing point (VP) detection: Hough-transform \cite{hough1} on Canny-edge \cite{canny} filtered images. The VP detection process is summarized in Algorithm \ref{alg:vp_detection}. Given a gray-scale image $x$ with height $H$ and width $W$, we do the following to estimate the VP:

\textbf{Pre-process}. To make edge detection more robust, morphological transform with opening (erosion followed by dilation) \cite{morph} is first used to denoise input images. Canny edge filter is then implemented to get the edge map. As the ``sky'' area constitutes the top part of images, we only do Canny edge filtering for the bottom $2/3$ part of images. Next, Hough-transform is applied to the edges, achieving a set containing the lines detected. For each line $\ell$ in the set, we denote its slope ($\Delta H / \Delta W$) as $\mathrm{slope}(\ell)$.

\textbf{Line selection}. Once having the Hough-lines \cite{hough1}, we decide which lines are to retain or discard. We design a criterion based on the likelihood of the lines that could be near the VP. On the one hand, as VPs are normally located around the center of the image, we delete the lines that are more than $d_{\rm max} = 160$ pixels away from the image center. Besides, we find that most horizontal and vertical lines (\eg, trees, wires) do not contribute to the VP detection. As a consequence, we delete any line $\ell$ from the Hough-line set that has $\mathrm{slope}(\ell)\notin S$, where $S = (-5, -0.2)\cup (0.2, 5)$ is a pre-defined slope acceptance interval.

\textbf{Cell vote}. After removing undesired lines, we compute the line intersections between line pairs. Upon obtaining $N_{\rm line}$ lines after line selection, we would get $N_{\rm line} (N_{\rm line} - 1) / 2$ intersections, notated as $R$. If the number of lines is too large, we randomly sample 100 lines among them. Next, we define several cells inside the image, where each cell is a rectangular box, and count the Hough-line intersections in it. In practice, we only parse cells in the lower part of the image, as the ``sky'' area takes the upper part of images. Finally, we choose the cell that includes the most intersections and return its center as the estimated VP position.

\begin{figure*}[t]
\centering
\includegraphics[scale=.55]{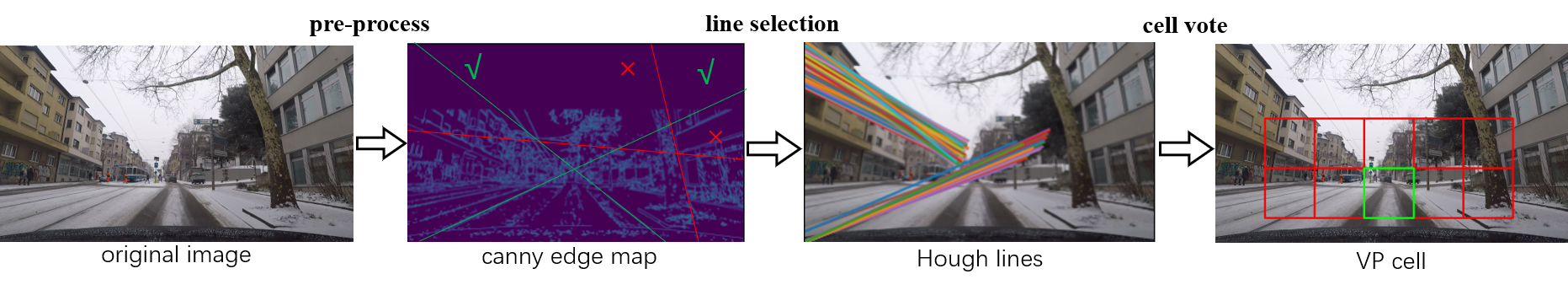}
\vspace{-5pt}
\caption{The VP detection pipeline. We first pre-process the input frame with morphology opening transform \cite{morph} and Canny edge filtering \cite{canny}. Hough-transform \cite{hough1} is then applied and lines that do not contribute to VP detection are discarded. Finally, cell vote is implemented to count the intersections in each cell to determine the final VP position.}
\label{fig:vp_detection}
\vspace{-5pt}
\end{figure*}

\begin{figure}[h!]
\centering
\begin{subfigure}{.245\textwidth}
  \centering
  \includegraphics[width=.9\linewidth]{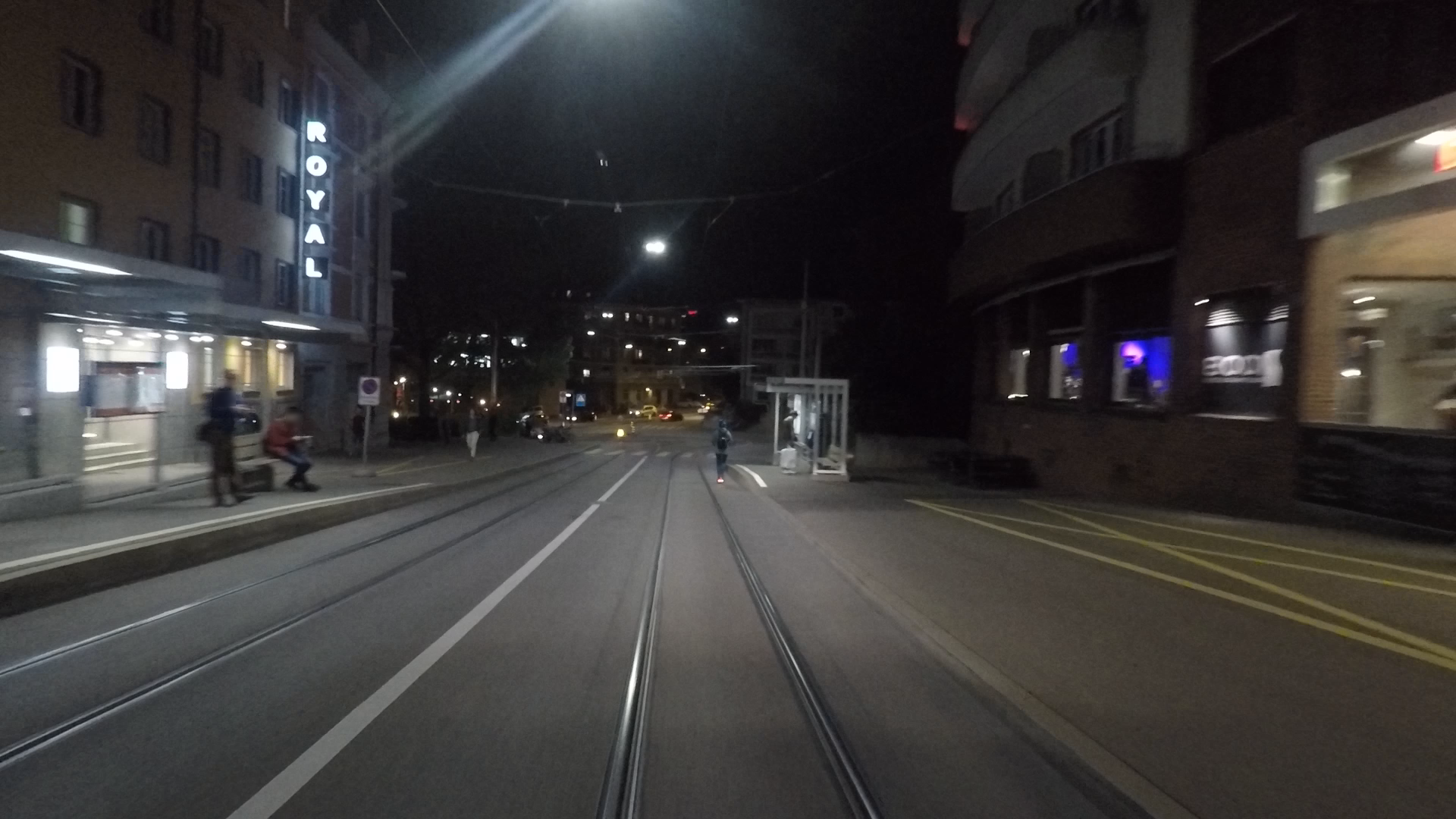}
  \caption{original image}
  \label{fig:sub1}
\end{subfigure}%
\begin{subfigure}{.245\textwidth}
  \centering
  \includegraphics[width=.9\linewidth]{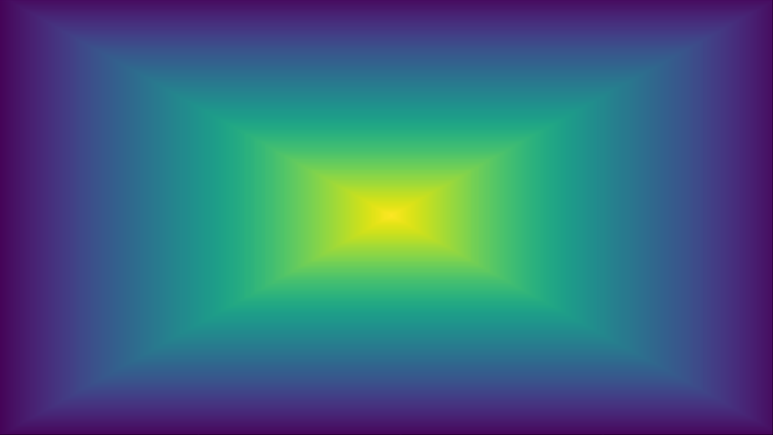}
  \caption{linear decreasing}
  \label{fig:sub2}
\end{subfigure}
\begin{subfigure}{.245\textwidth}
  \centering
  \includegraphics[width=.9\linewidth]{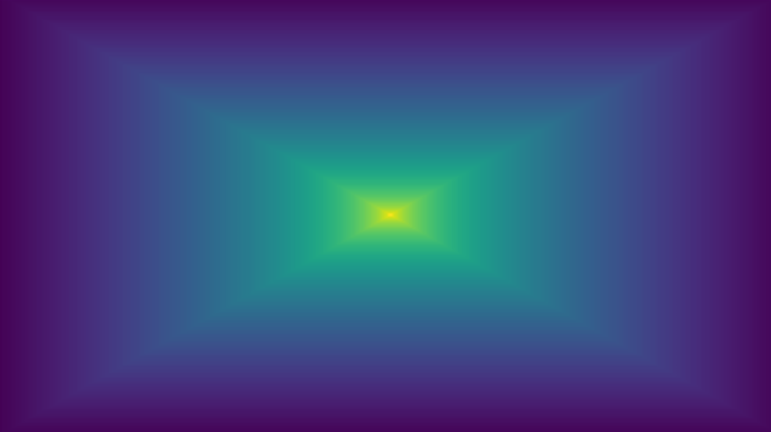}
  \caption{power decreasing}
  \label{fig:sub3}
\end{subfigure}%
\begin{subfigure}{.245\textwidth}
  \centering
  \includegraphics[width=.9\linewidth]{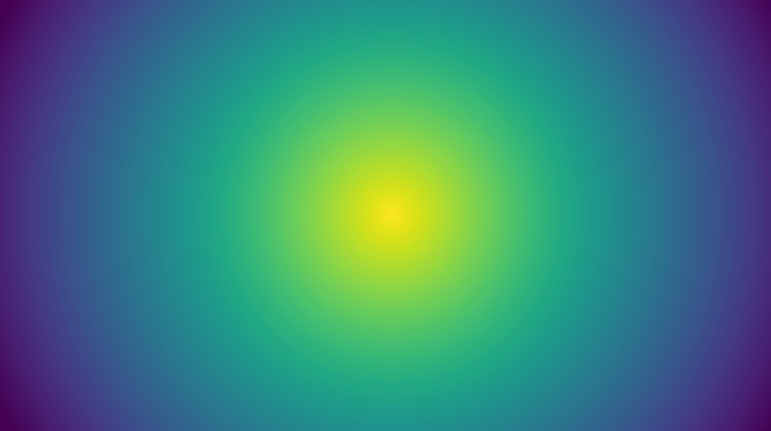}
  \caption{Euclidean decreasing}
  \label{fig:sub4}
\end{subfigure}

\caption{Different types of VP proximity map embeddings. (a) represents the input frame, (b) is the VP proximity map with linear decreasing. Compared with linear decreasing, the depth value in our power decreasing map (c) drops much faster around the VP. (d) denotes the proximity map with Euclidean decreasing, where the image aspect ratio $\frac{H}{W}$ is not considered.}
\label{fig:vp_maps}
\end{figure}

\begin{algorithm}
\caption{VP Detection}\label{alg:vp_detection}
\begin{algorithmic}[1]
\Require gray-scale image $x \in [0,255]^{H\times W}$, max central-to-line distance $d_{\rm max}$, slope acceptance interval $S$, square cells inside $x$ centered at $c_i = (H_i, W_i), i=1,...,N_{\rm cell}$ with size $L = \lfloor H/4 \rfloor$
\State $x \gets \mathrm{morphology\_opening}(x, \mathrm{kernel}=\mathbf{I}_{5\times5})$
\State $x \gets \mathrm{Canny\_edge}(x, 50, 150, \mathrm{apertureSize}=3)$
\State $\mathrm{lines} \gets \mathrm{Hough\_lines}(x, \rho = 1, \theta = \frac{\pi}{180}, \mathrm{thres} = 200)$
\State $c \gets (W/2, H/2)$
\For{$\ell \in \mathrm{lines}$}
\If{$d(c, \ell) > d_{\rm max}$ or $\mathrm{slope}(\ell)\notin S$}
    \State Delete $\ell$ from $\mathrm{lines}$
\EndIf
\EndFor
\State $R = \mathrm{find\_intersections}(\mathrm{lines})$
\For{$i=1,...,N_{\rm cell}$}
\State $n_i$ $\gets$ number hits of $R$ inside cell $i$
\EndFor
\State $i_{\rm opt} = \arg\max_{i} n_i$
\State \Return $c_{i_{\rm opt}}$
\end{algorithmic}
\end{algorithm}

After obtaining the VP, it is still a problem to pass the VP position to the model. As cropping operations are used in the pre-processing pipeline, we crop the VP proximity map along with the input frame. The cropped input frame and proximity map are concatenated as our new input.

The Hough-transform-based VP detection proves fast and robust in automated driving scenarios. However, challenges arise when dealing with images featuring messy or unclear edges. For instance, cross street scenes may exhibit multiple VPs, while crowded pedestrian areas can introduce noisy lines, affecting VP detection accuracy. To address these issues, we will combine the hand-crafted VP detection method with deep learning to strike a better balance between accuracy and inference speed in future works.

\section{Additional Ablation Studies}
\noindent \textbf{Effect of VP proximity embeddings}.\quad The linear VP proximity embedding is a VP-centered pseudo-depth map, where the depth of pixel $(x, y)$ is $1-\Delta D$, $\Delta D \propto \max \lbrace \frac{|y-\hat{y}^p_j|}{H}, \frac{|x-\hat{x}^p_j|}{W} \rbrace$ and $(\hat{x}^p_j, \hat{y}^p_j)$ is the VP pixel coordinate. Similarly, we introduce another two types of VP proximity maps: power and Euclidean decreasing (see \figref{fig:vp_maps}).
\begin{itemize}
    \item Linear (\figref{fig:sub2}): $\Delta D \propto \max \lbrace \frac{\Delta y}{H}, \frac{\Delta x}{W} \rbrace$

    In linear decreasing, the depth value of $(x, y)$ is linearly decreased according to its distance to the VP. It is fast to compute, and is our default option.
    
    \item Power (\figref{fig:sub3}): $\Delta D ^ 2 \propto \max \lbrace \frac{|\Delta y|}{H}, \frac{|\Delta x|}{W} \rbrace$

    In power decreasing, the square of the depth value is linearly decreased. It is more concentrated, but the depth drops faster around the VP.

    \item Euclidean (\figref{fig:sub4}): $\Delta D \propto \sqrt{(\frac{|\Delta y|}{H}) ^ 2 + (\frac{|\Delta x|}{H}) ^ 2}$

    In Euclidean decreasing, the depth value is linearly decreased according to the Euclidean distance. It is circular and isotropic, but ignores the image aspect ratio $\frac{H}{W}$.
    
\end{itemize}

We study the impact of above-mentioned VP proximity embeddings in \tabref{supp_embed}. Notably, \ourmodel{} with linear VP proximity embedding achieves the highest mIoU and mIA-IoU for ACDC \cite{acdc} and Cityscapes \cite{cityscapes}. The experiments with power and Euclidean embeddings perform slightly worse. The possible reason is that the Euclidean decreasing does not consider the image aspect ratio $\frac{H}{W}$. And the depth value of power decreasing drops too fast around the VP.

\noindent \textbf{Impact of the sampling coefficient $\Delta d$}.\quad We conduct experiments on different $\Delta d$ in \tabref{supp_delta_d}. We found that $\Delta d=1$ adequately covers fast-moving targets with good performance. MotionVP with $\Delta d=0$ only samples patches locally, which is unsuitable for high-speed driving scenarios and achieves worse mIoU and mIA-IoU. For $\Delta d > 1$, the performance of \ourmodel{} drops drastically, proving that larger $\Delta d$ is redundant for our VP-guided motion estimation.

\begin{table}[t!]
\centering
\footnotesize
  \renewcommand{\arraystretch}{.9}
  \setlength\tabcolsep{9pt}
\begin{tabular}{r|c|c|c}

\hline
 Embeddings  & mIoU (A.)$\uparrow$ & mIA-IoU (A.)$\uparrow$ & mIoU (C.)$\uparrow$ \\ \hline
  \rowcolor{mygray2}
  Linear           &    \textbf{77.48}  & \textbf{41.48}  & \textbf{82.46}\\
  Power          &    77.29   & 41.23  & 82.29 \\
  Euclidean       &    77.33   & 41.16   & 82.35  \\ 
\hline
\end{tabular}
\vspace{-5pt}
\caption{Ablation study of different VP proximity embeddings on ACDC (A.) and Cityscapes (C.) with MiT-B3 \cite{SegFormer} backbone.}
\label{supp_embed}
\end{table}

\begin{table}[t!]
\centering
\footnotesize
  \renewcommand{\arraystretch}{.9}
  \setlength\tabcolsep{12.5pt}
\begin{tabular}{r|c|c|c}
\hline
 $\Delta d$  & mIoU (A.)$\uparrow$ & mIA-IoU (A.)$\uparrow$ & mIoU (C.)$\uparrow$ \\ \hline
  0    &    76.74   & 40.57  & 81.83  \\
  \rowcolor{mygray2}
  1    &    \textbf{77.48}  & \textbf{41.48}  & \textbf{82.46}\\
  2    &    77.12   & 41.01  & 82.23 \\
  3    &    76.88   & 40.65   & 81.79  \\ 
\hline
\end{tabular}
\vspace{-5pt}
\caption{Ablation study of different sampling coefficients $\Delta d$ on ACDC (A.) and Cityscapes (C.) with MiT-B3 \cite{SegFormer} backbone.}
\vspace{-5pt}
\label{supp_delta_d}
\end{table}

\section{Detailed Pipelines}

\begin{figure*}
\centering
\begin{subfigure}{.9\textwidth}
  \centering
  \includegraphics[width=1.0\linewidth]{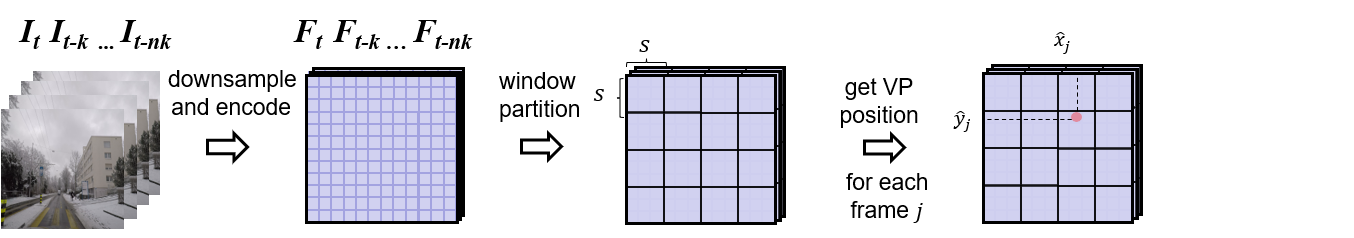}
  \captionsetup{width=1.0\linewidth}
  \caption{\textbf{Window partition and VP detection:} given $n+1$ input video frames, we first extract feature maps with pre-trained transformer encoder. The feature maps are then subdivided into feature blocks of size $s\times s$ (indexed by $i$). }
  \label{fig:vpmpve_sub1}
\end{subfigure}

\begin{subfigure}{1.0\textwidth}
  \centering
  \includegraphics[width=.95\linewidth]{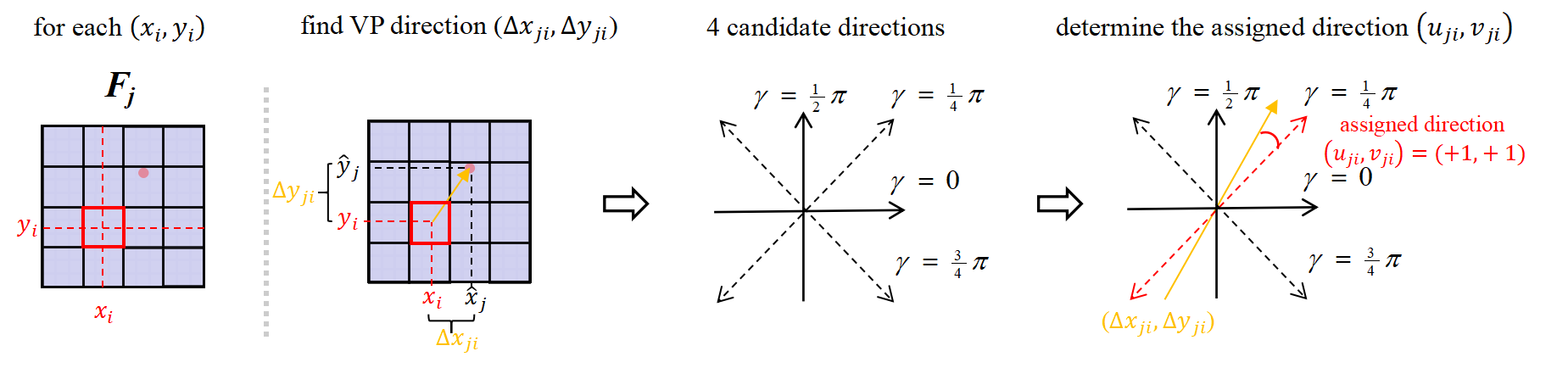}
  \captionsetup{width=0.9\linewidth}
  \caption{\textbf{Direction assignment:} we determine the assigned direction for each patch $(x_i, y_i)$. The assigned direction $(u_{ji}, v_{ji})$ is the closest candidate direction to vector $(\Delta x_{ji}, \Delta y_{ji})$, which points from the patch center to the VP. }
  \label{fig:vpmpve_sub2}
\end{subfigure}

\begin{subfigure}{1.0\textwidth}
  \centering
  \includegraphics[width=.96\linewidth]{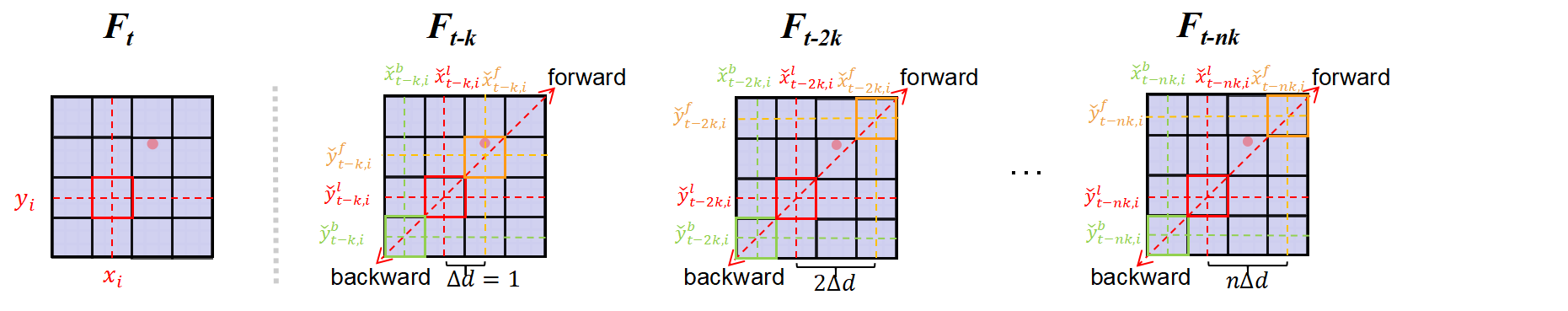}
  \captionsetup{width=0.9\linewidth}
  \caption{\textbf{Patch sampling:} after obtaining the assigned direction, we sample adjacent patches both forward and backward along the assigned direction. As the frame interval increases, the sampling distance also increases with the sampling coefficient $\Delta d$. But the sampled patches should not exceed the boundaries of the feature map.}
  \label{fig:vpmpve_sub3}
\end{subfigure}

\begin{subfigure}{1.0\textwidth}
  \centering
  \includegraphics[width=.95\linewidth]{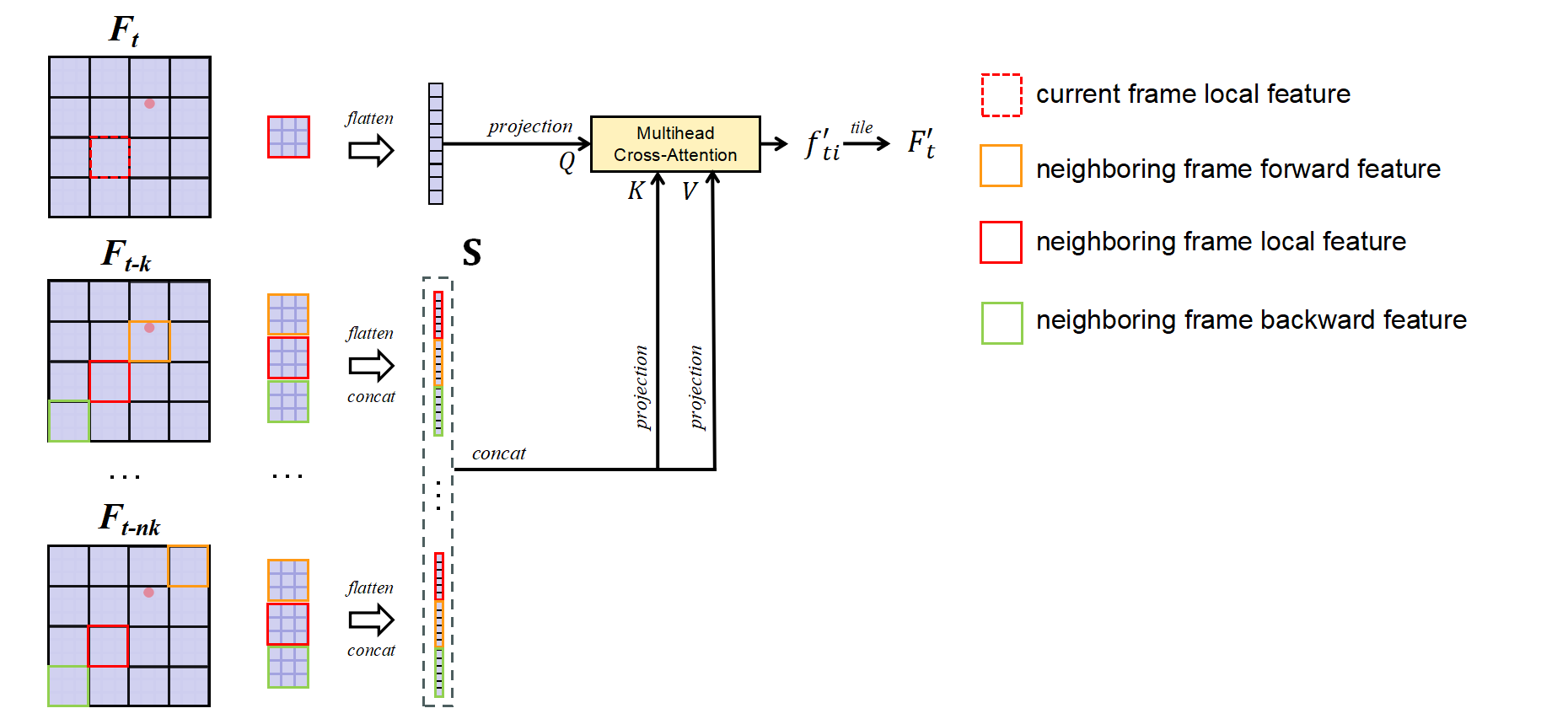}
  \captionsetup{width=0.9\linewidth}
  \caption{\textbf{Feature aggregation:} we generate dynamic context $F_t'$ with cross-attention \cite{attentionisallyouneed} operations. Specifically, for each patch $(x_i, y_i)$, the patch features of the current frame serve as queries, while the sampled features in neighboring frames serve as keys and values. After cross-attention, we achieve patch-level dynamic features $f_t'$, which are then simply tiled together to reconstruct the complete frame-level dynamic context $F_t'$.}
  \label{fig:vpmpve_sub4}
\end{subfigure}

\caption{Detailed MotionVP pipeline.}
\label{fig:vp_move}
\end{figure*}

\begin{figure*}
\centering
  \centering
  \includegraphics[width=.91\linewidth]{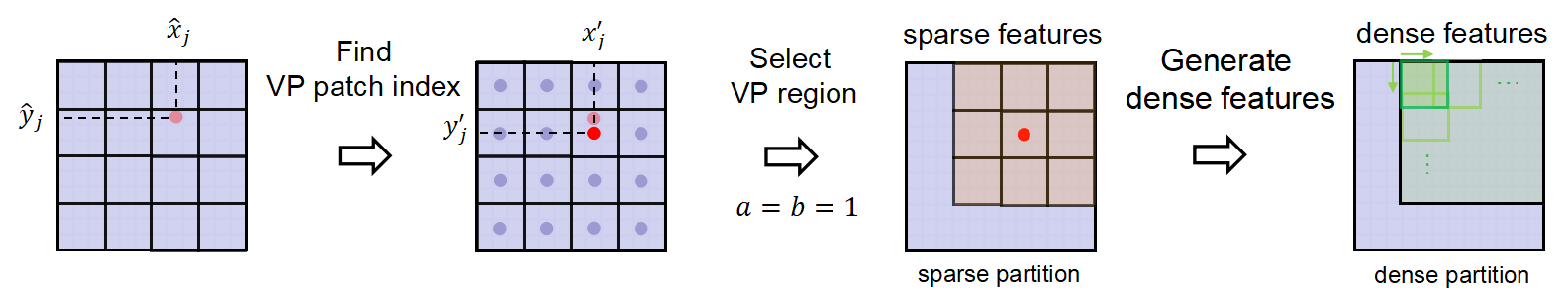}
\vspace{-10pt}
\caption{Detailed DenseVP pipeline. \textbf{Find VP patch index:} we find the closest patch to the VP as our VP patch. \textbf{Select VP region:} we select a rectangular region around the VP patch as our VP region $\mathbf{A}$. \textbf{Generate dense features:} the overlapping dense partition strategy is applied in the VP region, obtaining dense features $f_{\mathbf{A}}$.}
\label{fig:vp_dense}
\vspace{-2pt}
\end{figure*}

To exploit dynamic and static VP priors, we proposed MotionVP and DenseVP. MotionVP extracts dynamic context and can be divided into four parts: window partition and VP detection, direction assignment, patch sampling, and feature aggregation. DenseVP augments the dynamic context with finer attention around the VP region and consists of three steps: find VP patch index, select VP region, and generate dense features. The augmented dynamic context is sent to the prediction head for the final prediction. The details of MotionVP and DenseVP pipelines are shown in \figref{fig:vp_move} and \figref{fig:vp_dense}, while \tabref{tab:Notations} explains types, domains, and meanings of the symbols from MotionVP and DenseVP.

\begin{table*}[t!]
  \centering
  \renewcommand{\arraystretch}{1.0}
  \renewcommand{\tabcolsep}{7.5pt}
  \caption{Table of symbols, their types, domains, and meanings.
  }\label{tab:Notations}
  \vspace{-8pt}
  \begin{tabular}{c|c|c|c|l} \hline \hline
    Symbol & Type & Size (length) & Domain & Meaning \\ \hline
    $\mathcal{I}$ & set & $n+1$ & -  & a set of input frames
    \\ \hline
    $\mathcal{T}$ & set & $n+1$ & -   & a set of timestamps
    \\ \hline
    $\mathcal{F}$ & set & $n+1$ & -   & a set of feature maps
    \\ \hline
    $\mathcal{D}$ & set & $n+1$ & - & a set of patch indexes
    \\ \hline
    $\mathcal{A}$ & set & $(2a+1)(2b+1)$ & - & a set of sparse patch indexes of the VP region
    \\ \hline
    $\mathcal{V}$ & set & $4$ & - & a set of vector representations of candidate directions
    \\ \hline
    $\mathcal{S}$ & set & $3n$ & - & a set of sampled features in $n$ neighboring frames
    \\ \hline  \hline

    $c$ & scalar & - & $\mathbb{N}$  & number of feature channels\\ \hline
    $h, w$ & scalar & - & $\mathbb{N}$  & spatial height/width of the feature map\\ \hline
    $H, W$ & scalar & - & $\mathbb{N}$  & spatial height/width of the input frame\\ \hline
    $k$ & scalar & - & $\mathbb{N}$ & frame sampling interval \\ \hline
    $K$ & scalar & - & $\mathbb{N}$ & number of semantic classes \\ \hline
    $s$ & scalar & - & $\mathbb{N}$  & size of the feature block\\ \hline 
    $m$ & scalar & - & $\mathbb{N}$  & number of dense patches in the VP region\\ \hline 
    $\Delta d$ & scalar & - & $\mathbb{N}$  &
    sampling coefficient \\ \hline  
    \hline
    $\left ( \hat{x}_j , \hat{y}_j \right ) $ & coordinate & $2 \times 1$ & $\mathbb{R}$  & patch-level VP position in frame $j$ \\ \hline
    $\left ( \hat{x}_j^p , \hat{y}_j^p \right ) $ & coordinate & $2 \times 1$ & $\mathbb{N}$  & pixel-level VP position in frame $j$ \\ \hline
    $(x_{i}, y_{i})$ & coordinate & $2 \times 1$ & $\mathbb{N}$   & index of the $i$-th patch
    \\ \hline
    $(\check{x}_{ji}^{f}, \check{y}_{ji}^{f})$ & coordinate & $2 \times 1$ & $\mathbb{N}$  & forward sampled patch index for the $i$-th patch in frame $j$\\ \hline
    $(\check{x}_{ji}^{b}, \check{y}_{ji}^{b})$ & coordinate & $2 \times 1$ & $\mathbb{N}$  & backward sampled patch index for the $i$-th patch in frame $j$\\ \hline
    $(\check{x}_{ji}^{l}, \check{y}_{ji}^{l})$ & coordinate & $2 \times 1$ & $\mathbb{N}$  & locally sampled patch index for the $i$-th patch in frame $j$ \\ \hline 
    $(x'_{j}, y'_{j})$ & coordinate & $2 \times 1$ & $\mathbb{N}$   & VP patch index in frame $j$
    \\ \hline \hline

    $I_t$ & matrix & $H\times W$  & $\mathbb{R}$
    & frame in time $t$ \\ \hline
    
    $F_t$ & matrix & $c \times h \times w$ & $\mathbb{R}$ &
    feature map for frame $t$ \\ \hline

    $f_{ti}$ & matrix & $c \times s^2$ & $\mathbb{R}$ &
    patch-level feature for the $i$-th patch in frame $t$ \\ \hline
    
    $F_{tl}, F_{th}$ & matrix & $c \times h \times w$ & $\mathbb{R}$ &
    low/high-resolution feature map of $I_t$ \\ \hline

    $\check{f}_{ji}$ & matrix & $c \times 3s^2$ & $\mathbb{R}$ & sampled features for the $i$-th patch in frame $j$\\ \hline
    
    ${F}'_{t}$ & matrix & $c \times h \times w$ & $\mathbb{R}$  &
    frame-level dynamic features in frame $t$\\ \hline

    ${f}'_{ti}$ & matrix & $c \times s^2$ & $\mathbb{R}$  &
    patch-level dynamic features for the $i$-th patch in frame $t$\\ \hline

    ${f}_{\mathcal{A}}$ & matrix & $c \times ms^2$ & $\mathbb{R}$ & dense features of VP region \\ \hline

    ${F}''_{t}$ & matrix & $c \times h \times w$ & $\mathbb{R}$  &
    augmented dynamic context in frame $t$\\ \hline

    ${E}$ & matrix & $h \times w$ & $\mathbb{R}$  &
    VP proximity map\\ \hline
    
    ${Q}, {Q}_{c}$ & matrix & $c \times K$ & $\mathbb{R}$  &
    learnable/contextualized queries in CMA\\ \hline

    ${F}_{m}$ & matrix & $c \times K$ & $\mathbb{R}$  &
     the merged context \\ \hline

    $G_{nz}$ & matrix & $H\times W$ & $\left \{ 0, 1 \right \}$  & ground truth of the $z$-th class in the $n$-th image\\ \hline

    $P_{nz}$ & matrix & $H\times W$ & $\left \{ 0, 1 \right \}$  & prediction of the $z$-th class in the $n$-th image\\ \hline
    
    $M_n$ & matrix & $H\times W$ & $\left \{ 0, 1 \right \}$  & invalid mask of the $n$-th image\\ \hline
    \hline

  \end{tabular}
\end{table*}

\end{document}